\documentclass[sigconf]{acmart}

\usepackage{algorithm}
\usepackage{algorithmic}

\usepackage{amsmath}
\usepackage{tipa}
\usepackage{colortbl}
\usepackage{pifont} 
\usepackage{makecell}
\usepackage[xcdraw,table]{xcolor}
\usepackage{float} 
\usepackage{caption}
\usepackage{subcaption}
\usepackage{booktabs}
\usepackage{graphicx}
\usepackage{multirow}
\usepackage{array}
\usepackage{algorithm}
\usepackage{algorithmic}

\usepackage{booktabs}
\usepackage{tabularx}
\usepackage{multirow}
\usepackage{placeins}

\newcommand{\cmark}{\ding{51}}
 
\newcommand*{\boldcheckmark}{\textbf{\cmark}}

\renewcommand{\arraystretch}{1.2}

\definecolor{Gray}{gray}{0.95}
\definecolor{darkgreen}{rgb}{0.0, 0.5, 0.0}
\usepackage{tikz}

\makeatletter

\newcommand{\eqtag}{%
  \refstepcounter{equation}%
  \tag{\tagform@{\theequation}}%
}

\newcommand{\condtag}[1]{%
  \refstepcounter{equation}%
  \tag{\text{[#1]}\quad \tagform@{\theequation}}%
}
\makeatother

\newcommand{\vx}{\boldsymbol{x}}

\AtBeginDocument{%
  }

\copyrightyear{2026}
\acmYear{2026}
\setcopyright{cc}
\setcctype{by}
\acmConference[WWW '26] {Proceedings of the ACM Web Conference 2026}{April 13--17, 2026}{Dubai, United Arab Emirates.}
\acmBooktitle{Proceedings of the ACM Web Conference 2026 (WWW '26), April 13--17, 2026, Dubai, United Arab Emirates}
\acmISBN{979-8-4007-2307-0/2026/04}
\acmDOI{10.1145/3774904.3792411}

\newcolumntype{Y}{>{\centering\arraybackslash}X}
\newcolumntype{L}{>{\raggedright\arraybackslash}X}
\newcolumntype{C}{>{\centering\arraybackslash}X}

\begin{document}

\title{BiasEdit: A Training-Free Bias-Detect-and-Edit Framework for Learning Fair Visual Classifiers}

\author{Jungwook Seo}
\email{zungwooker@hanyang.ac.kr}
\orcid{0009-0005-8335-5898}
\affiliation{%
  \institution{Hanyang University}
  \department{Department of Artificial Intelligence}
  \department{BAIK Lab}
  \city{Seoul}
  \country{South Korea}
}

\author{Yoonsik Park}
\email{yoons1595@hanyang.ac.kr}
\orcid{0009-0007-4975-5294}
\affiliation{%
  \institution{Hanyang University}
  \department{Department of Data Science}
  \department{BAIK Lab}
  \city{Seoul}
  \country{South Korea}
}

\author{Changmin Lee}
\email{dlckdals0204@hanyang.ac.kr}
\orcid{0009-0003-0539-407X}
\affiliation{%
  \institution{Hanyang University}
  \department{Department of Data Science}
  \department{BAIK Lab}
  \city{Seoul}
  \country{South Korea}
}

\author{Sungyong Baik$^{\dagger}$}
\thanks{$^{\dagger}$Corresponding author.}
\email{dsybaik@hanyang.ac.kr}
\orcid{0000-0001-5702-4618}
\affiliation{%
  \institution{Hanyang University}
  \department{Department of Data Science}
  \department{Department of Artificial Intelligence}
  \department{BAIK Lab}
  \city{Seoul}
  \country{South Korea}
}

\renewcommand{\shortauthors}{Jungwook Seo et al.}

\begin{abstract}
Visual data from the Web power image classifiers, which underpin web services, including recommendation and moderation. However, Web data contain spurious correlations and social biases, and neural networks tend to learn biases in data. This can reinforce unfairness in web services and the web data, leading to a vicious cycle. In image classification, networks learn bias attributes for a specific class when most images contain the same attribute for a given class. Hence, training a fair and debiased classifier from a biased dataset demands handling imbalance between a majority of images with bias attributes (bias-aligned samples) and a minority without (bias-conflict samples). In this work, we introduce BiasEdit, a modular framework that automatically detects bias attributes from the original dataset and edits them to construct a debiased dataset. Specifically, BiasEdit first detects unknown bias attributes via statistical dependence and mutual information analysis of visual–linguistic representations, and then explicitly edits those attributes using text-guided image editing to generate realistic bias-conflict samples. Unlike prior works that assume known bias attributes or rely on synthetic mixing, our method operates without manual annotations and leverages off-the-shelf vision–language and editing models. BiasEdit addresses a fundamental challenge in Web-sourced visual AI, mitigating dataset-induced bias and achieving state-of-the-art debiasing performance even when training data are fully biased.
\end{abstract}




\maketitle
\section{Introduction}
The Web has become the largest and most diverse source of visual data for training modern AI systems~\cite{schuhmann2022laion, deng2009imagenet}. 
From social media platforms and e-commerce sites to search and content moderation pipelines, Web-sourced visual data power classifiers that interpret and organize online content~\cite{he2016deep, hu2018web, covington2016deep, wu2017automatic, xu2022rethinking, kiela2020hateful}. 
However, such data inevitably reflect the social and contextual biases present in human behavior—for example, correlations between gender and occupation or between background and object category. 
Neural networks trained on these datasets tend to exploit these spurious correlations, resulting in biased classifiers, undermining its fidelity and fairness~\cite{aldahoul2025ai, geirhos2020shortcut}.
When deployed in Web services, these models can reinforce stereotypes, distort recommendations, or unfairly filter content, amplifying existing biases at scale.

\begin{figure}[t!]
    \centering
    \includegraphics[width=\linewidth]{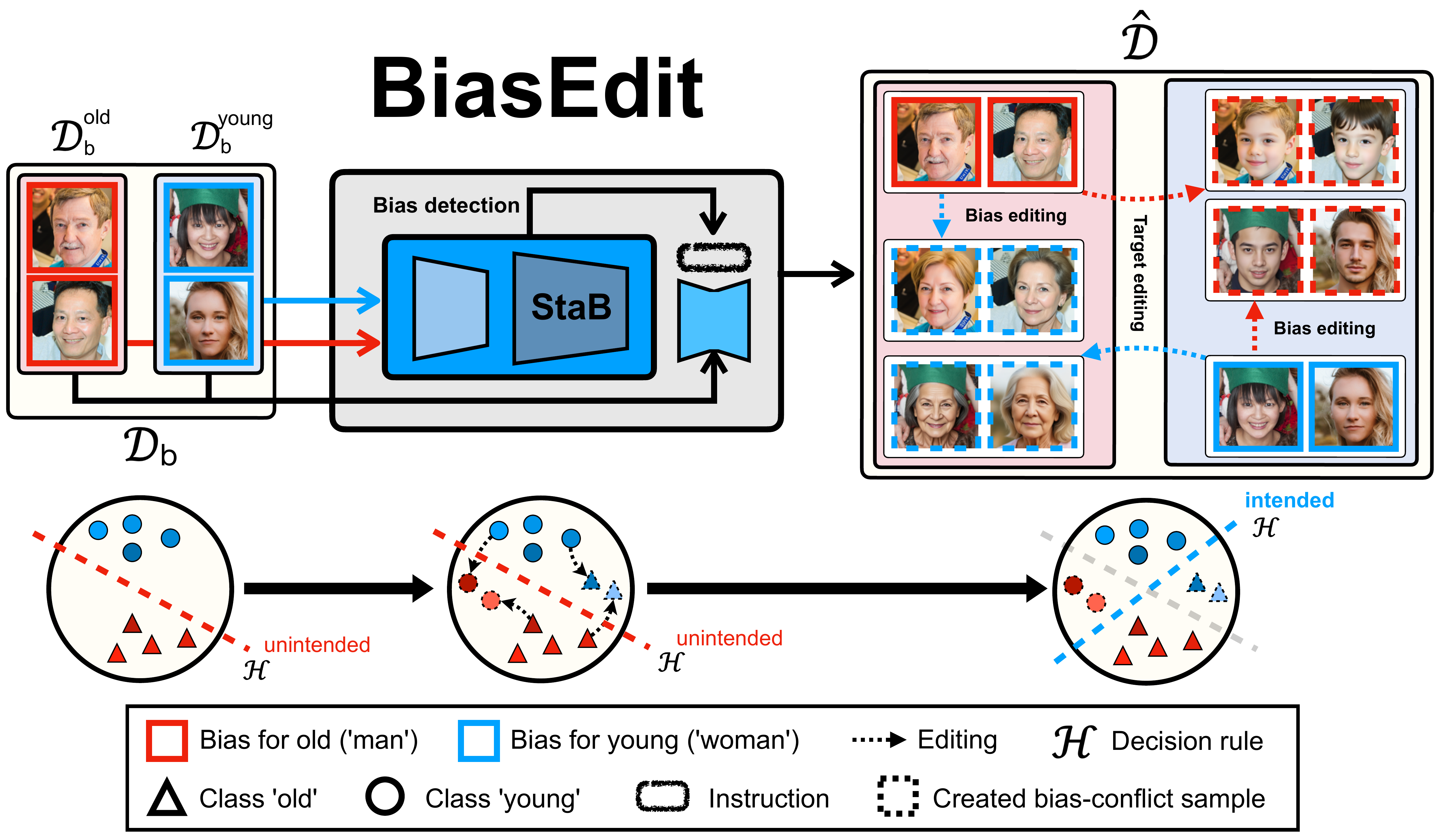}
    \vspace{-0.4cm}
    \caption{
    \textbf{Illustration of the BiasEdit pipeline.} 
    BiasEdit first detects unknown biases.
    Then, detected biases are used to create bias-conflict samples via editing of bias and target attributes, such that dataset biases will be mitigated.
    }
    \vspace{-0.5cm}
    \label{fig:head}
\end{figure}

One of the causes for the problem is the tendency of neural networks to learn to exploit bias attributes that exist in datasets~\cite{geirhos2018imagenet}.
A bias attribute refers to an attribute that frequently appears in a given dataset and forms a spurious correlation (i.e., non-causal misleading relationship) with the target attribute. 
For example, if a large number of `pedestrian' images in a dataset contain `crosswalk' background, a spurious correlation is present between a bias attribute (`crosswalk') and a target attribute (`pedestrian').
The abundance of bias attributes in a biased dataset makes it easier for neural networks to minimize empirical risks by learning the spurious correlations between bias and target attributes, a phenomenon known as shortcut learning~\cite{geirhos2020shortcut}.

Debiasing aims to train models to focus on target features—essential characteristics that define each class—while ignoring spurious correlations in biased datasets. Early approaches~\cite{sagawa2019distributionally,kim2019learning,tartaglione2021end,qraitem2023bias} assume that all bias attributes are known, resulting in methods tailored to these biases. 
Such strong assumption requires extensive attribute annotation, limiting scalability and generalizability to unknown or overlooked biases. 
Some works leverage prior bias knowledge~\cite{wang2019learning,bahng2020learning}, but access to such information is scarce in real-world settings.

More recently, research has advanced debiasing without the known-bias assumption~\cite{nam2020learning,lee2023revisiting,park2024enhancing,jung2024simple}. These methods typically identify bias-conflict samples—based on the notion that bias attributes are learned faster than target attributes—and upweight them during training. However, they are limited by the small number of bias-conflict samples, which poses a risk of overfitting.

To overcome overfitting, strategies for generating bias-conflict samples have been explored, including generative models~\cite{goodfellow2014generative,jung2023fighting,kim2021biaswap}, mixup augmentation~\cite{zhang2017mixup,hwang2022selecmix}, or adversarial examples~\cite{lim2023biasadv}.
Nevertheless, these generated samples are often unrealistic and retain substantial bias, as most methods do not explicitly detect or manipulate bias attributes but use indirect approaches like sample mixing or minor pixel perturbations.

In this work, by contrast, we aim to create realistic bias-conflict samples.
To this end, we propose a framework that performs three steps: (1) \textit{detecting} bias attributes; (2) \textit{editing} both bias attributes and target attributes in order to produce realistic bias-conflict samples; and (3) train a \textit{debias}ed classifier on newly constructed datasets, as illustrated in Figure~\ref{fig:head}.
To detect bias attributes, we first extract attributes from images using a vision-language model~\cite{huang2023tag2text, liu2024improved}.
Then, we propose a statistical bias attribute detection to identify bias attributes from extracted attributes by using statistical relationships across classes, specifically statistical dependence and mutual information.
In order to gain fine control over the editing of bias and target attribute in newly created images, we employ text-guided editing~\cite{liu2025step1x, brooks2023instructpix2pix, zhang2024magicbrush} to explicitly edit bias and target attributes in images, thereby producing bias-conflict samples.
Experimental results demonstrate that our proposed approach, dubbed \textbf{BiasEdit}, not only creates realistic bias-conflict samples but also outperforms existing baselines.
Notably, BiasEdit achieves robust performance even in extreme scenarios, such as when the dataset contains up to $100\%$ bias-aligned samples.
\vspace{-0.05cm}
\section{Related works}
The growing reliance on Web-sourced visual data has amplified long-standing concerns about bias in machine learning. 
Web-collected datasets inevitably contain social and contextual imbalances that reflect human behavior and online content patterns.
These biases propagate through downstream classifiers, affecting fairness in Web services such as content moderation, search, and recommendation. 
Consequently, debiasing techniques have become a critical component of responsible Web AI.
Early debiasing approaches assume the full knowledge of bias attributes, relying on meticulously labeled datasets or known bias types~\cite{sagawa2019distributionally,kim2019learning,tartaglione2021end,qraitem2023bias,geirhos2018imagenet,wang2019learning,li2020shape,bahng2020learning}. 
This assumption, however, limits their applicability in real-world scenarios where such detailed annotation is costly and rarely available.

To address this limitation, recent studies tackle unknown-bias scenarios.
Many leverage the observation that models learn bias attributes faster than target attributes, using loss from an auxiliary biased model to identify and up-weight bias-conflict samples~\cite{nam2020learning,lee2023revisiting,jung2024simple}.
Others utilize attribute representations~\cite{lee2021learning,li2023partition} or analyze relationships between bias attributes and network depth~\cite{wang2024navigate,sreelatha2024denetdm}.

We note that a fundamental challenge is the severe imbalance: bias-conflict samples are often scarce. 
To overcome this, some works generate bias-conflict samples using simple augmentations or manipulations, such as mixup, interpolation, adversarial attacks, or feature-space augmentation~\cite{hwang2022selecmix,zhang2023learning,lim2023biasadv,lee2021learning}; however, these usually fail to produce sufficiently realistic or diverse examples.

Some studies~\cite{kim2021biaswap,lee2023revisiting,jung2024simple} identify bias-aligned or conflict samples under unknown-bias settings, but do not directly detect the bias attribute, facing similar limitations.
Approaches discovering bias-opposite concepts~\cite{kim2024discovering} are also limited, as finding opposite concepts does not equate to identifying the actual bias attribute.

In parallel, research on fair image generation—while not focused on discriminative models—has explored techniques such as latent perturbations, learnable prompts, and retrieval-augmented generation to achieve fairness~\cite{ramaswamy2021fair,zhang2023iti,shrestha2024fairrag}. 
These methods, however, require additional training procedures of generative models and curated reference samples.
In contrast, our flexible framework leverages an off-the-shelf pretrained model and focuses the editing primarily on the necessary components (i.e., bias attributes), facilitating the creation of realistic bias-conflict samples without additional generative-model training or curated reference samples.
\begin{figure*}[ht!]
    \centering
    \includegraphics[width=\linewidth]{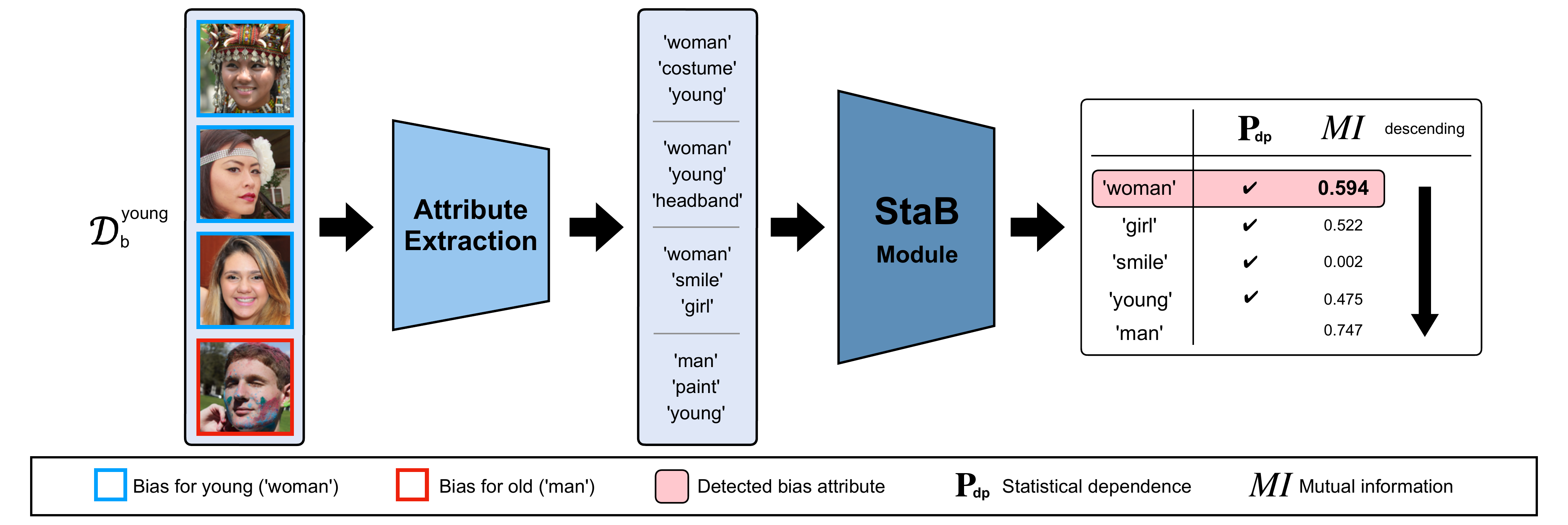}
    \vspace{-0.7cm}
    \caption{\textbf{Illustration of the process of detecting bias using the StaB module in BiasEdit.} 
    BiasEdit first statistically detects biases for each class based on the attributes identified in the dataset.
    The attribute that is detected with high mutual information, while satisfying the statistical dependence condition is considered the bias. 
    The above figure illustrates the process of detecting bias in the BFFHQ dataset.
    }
    \label{fig:main1}
    \vspace{-0.2cm}
\end{figure*}

\vspace{-0.2cm}
\section{Proposed method}
\noindent\textbf{Problem formulation.}
Given a biased train set $\mathcal{D}_b=\{(\vx_i, y_i)\}^{N}_{i=1}$, we aim for debiasing, where we train a debiased classifier $f_d: \mathcal{X} \rightarrow \mathcal{Y}$ that correctly classifies an input sample $\vx \in \mathcal{X}$ as its corresponding class label $y \in \mathcal{Y}$, which is one of $\mathcal{C}$ classes present in the dataset.
In a biased dataset, a majority of input data contain an attribute $a\in\mathcal{A}$ for a certain class $c$, leading to spurious correlations between them, which is not a causal relationship.
We call such attribute a bias attribute  $a^c_\text{bias}$ for a class $c\in\mathcal{C}$. 
We refer to an example of class $c$ as a bias-aligned sample $\vx_\text{ba}^c$, if it contains a bias attribute $a^c_\text{bias}$ and a bias-conflict sample $\vx_\text{bc}^c$ otherwise.
Standard training of a classifier $f$ via empirical risk minimization (ERM) with a biased dataset $\mathcal{D}_b$ will result in a biased classifier $f_b$.
A biased classifier $f_b$ makes use of these easy-to-learn bias attributes $a^c_\text{bias}$, rather than target attributes of a class, thereby learning unintended decision rules.

\noindent\textbf{Overall framework.}
To achieve debiasing under realistic scenarios, we assume no knowledge of bias attributes present in a given biased dataset $\mathcal{D}_b$.
A biased dataset inherently introduces a data-imbalance problem, as a majority of samples for a class $c$ will be bias-aligned samples $\vx_\text{ba}^c$, with only few samples being bias-conflict samples $\vx_\text{bc}^c$.
In this work, we aim to directly resolve the lack-of-diversity and data-scarcity issues by creating bias-conflict samples.
To this end, we first (1) statistically detect bias attributes, then (2) use the detected bias attributes to produce bias-conflict samples by directly editing attributes, and finally (3) construct a bias-reduced dataset to train a debiased classifier.

\subsection{Bias attribute detection}
\label{sec:StaB}
\noindent\textbf{Detecting visual-linguistic attributes.}
To detect unknown bias attributes, we begin by attempting to identify all attributes present in the dataset. 
In this work, we assume that such attributes can be regarded as keywords that are readily available at large scale from the internet.
Specifically, we use a vision-language model (VLM)~\cite{liu2024improved} to extract attribute keywords from each sample.
For example, when a sample of a male doctor is input, the VLM outputs keywords such as \textit{\{`male', `tie', `glasses'\}}.
We denote this extraction function as $T$, where for an input sample $\vx$, $T(\vx)$ outputs a set of attribute keywords that represent the contents of the sample.

Using the VLM, we extract a set of attributes from each sample in the biased dataset $\mathcal{D}_b$:
\begin{align}
\mathcal{A}^{c}_{j} &= \{a \mid a \in T(\vx^c_j)\},\label{eq:extract-attrubute}\\
\mathcal{A}^{c} &= \bigcup_{j} \mathcal{A}^{c}_{j},
\end{align}
where $c$ denotes a class of a sample, while $j$ denotes the index of a sample $\vx$ belonging to class $c$. Here, $\mathcal{A}^{c}_{j}$ is the set of attributes extracted in the $j$-th sample of class $c$.

For statistical bias attribute detection, we first compute the occurrence count $n^{c}(a)$ of each detected attribute $a \in \mathcal{A}^{c}$ for each class $c$:
\begin{equation}
    n^{c}(a) = \sum_{j=1}^{N_c} \mathbb{I}\big(a \in \mathcal{A}^{c}_{j}\big),
\end{equation}
where $N_c$ denotes the number of samples in class $c$, and $\mathbb{I}(\cdot)$ is the indicator function defined as
\[
  \mathbb{I}(\text{condition}) = 
  \begin{cases}
    1, & \text{if the condition holds}, \\
    0, & \text{otherwise}.
  \end{cases}
\]

\begin{figure*}[ht!]
    \centering
    \includegraphics[width=\linewidth]{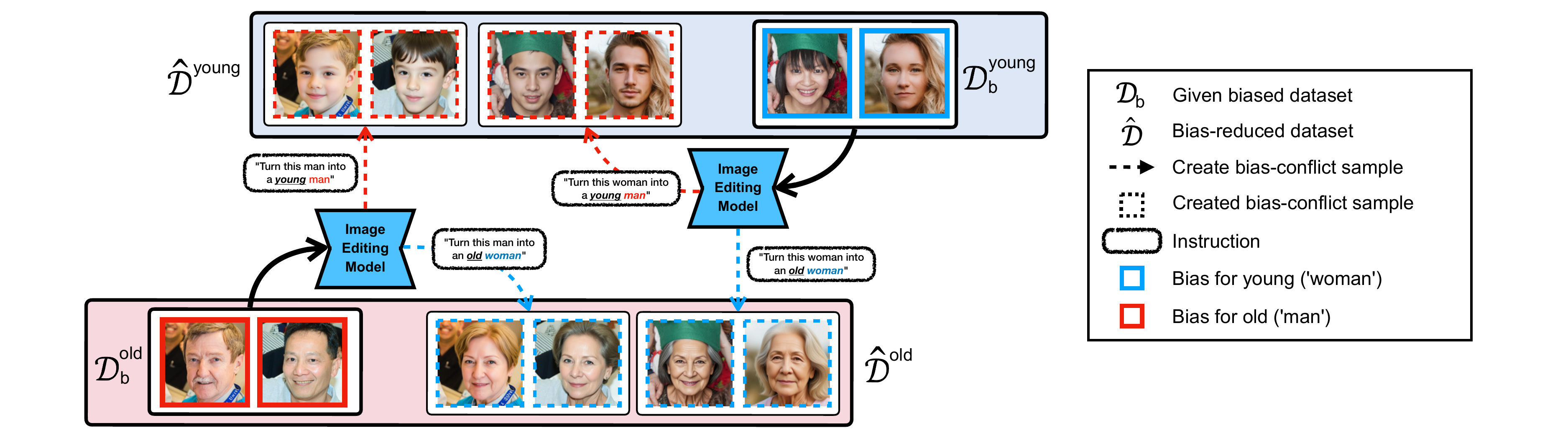}
    \vspace{-0.5cm}
    \caption{\textbf{Illustration of producing bias-conflict samples by editing the bias and target attributes from a biased dataset.}
    BiasEdit produces bias-conflict samples by directly editing the bias and target attributes from a biased dataset using the biases detected for each class by the StaB module, while aiming to preserve other attributes. The produced bias-conflict samples can help mitigate spurious correlations in the dataset.
    }
    \label{fig:main2}
    \vspace{-0.3cm}
\end{figure*}

\noindent\textbf{Statistical bias attribute detection.}
In order to identify bias attributes from the extracted attributes, we introduce StaB (Statistical Bias Detector), a module designed to statistically determine whether each extracted attribute constitutes a bias in a given dataset.
StaB aims to identify which of the extracted attributes most strongly induces spurious correlations with class labels.
To this end, StaB evaluates whether a given attribute meets the two conditions:
\begin{enumerate}
    \item \textbf{Condition 1. Statistical Dependence.} A bias attribute and a class label are statistically dependent.
    \item \textbf{Condition 2. Mutual Information.} Given a bias attribute, the uncertainty (entropy) of the class label decreases.
\end{enumerate}

Condition 1 is based on the definition of bias~\cite{barocas2023fairness}.
A given attribute $a$ and a class label $c$ are considered to be dependent if the probability of class $c$ increases when conditioned on a given attribute $a$.
However, Condition 1 may be vulnerable to rare attributes, not necessarily representing spurious correlations.  
For example, if there is only one sample with a `snow' attribute, the `snow' attribute will satisfy Condition 1, even though it provides little information about its corresponding class. 

Condition 2 interprets the bias issue in classification tasks from the perspective of information uncertainty.  
For an attribute to be a bias attribute and affect learning, an attribute and a class should form strong spurious correlation: much information about a class should be obtained after observing an attribute.  
To quantify such information gain of a class conditioned on an attribute, we calculate mutual information $\text{MI}(Z_c; W_a)$, where $Y$ denotes the class label, $Z_c=\mathbb{I}(Y=c)$, and $W_a=\mathbb{I}(a \in T(X))$.
Specifically, $\text{MI}(Z_c; W_a) = H(Z_c)-H(Z_c|W_a)$, where $H$ denotes entropy.
The above process can be summarized as follows:
\begin{gather}
\mathbf{P}_{\mathrm{dp}} := \mathbb{I}\left( P(Z_c=1|W_a=1) > P(Z_c=1) \right), \quad \text{[Condition 1]} \label{eq:condition1} \\[0.8em]
\mathcal{A}^c_{\mathrm{dp}} = \left\{ a \in \mathcal{A}^c : \mathbf{P}_{\mathrm{dp}} = 1 \right\}, \label{eq:condition1a} \\[0.4em]
a^c_{\mathrm{bias}} = \arg\max_{a \in \mathcal{A}^c_{\mathrm{dp}}} \text{MI}(Z_c; W_a),  \quad\text{[Condition 2]} \label{eq:condition2} \\[0.4em]
\mathcal{B} = \left\{ a^c_{\mathrm{bias}} \mid c \in \mathcal{C} \right\}. \label{eq:bias_attr}
\end{gather}

Here, \(\mathbf{P}_{\mathrm{dp}}\) denotes an indicator variable representing whether Condition 1 is satisfied.
\(\mathcal{A}^c_{\mathrm{dep}}\) denotes the set of attributes for class \(c\) that satisfy \(\mathbf{P}_{\mathrm{dp}}\), and \(\mathcal{B}\) denotes the set of identified bias attributes for each class.
To elaborate on Equation~\ref{eq:condition1}, $P(Z_c=1|W_a=1)$ represents the probability of class $c$ given attribute $a$, defined as the ratio $P(Z_c=1|W_a=1) = \frac{n_c(a)}{\sum^{|C|}_{l=1} n_l(a)}$.
Similarly, the marginal probability $P(Z_c=1)=N_c/N$ captures the overall proportion of samples belonging to class $c$ within a dataset.
A strong dependence of class $c$ on attribute $a$ (Condition 1) means that the distribution of attribute $a$ is skewed toward class $c$.
The overall procedure for detecting bias attributes using StaB is illustrated in Figure~\ref{fig:main1}.
We remove an attribute corresponding to the class label itself when determining the bias attribute; this process remains cost-effective, as the target label is available in supervised training.
Through the process, we select the attribute with the highest $\text{MI}(Z_c;W_a)$ as the bias attribute for each class by dependence.

\subsection{Producing bias-conflict samples by editing biases and targets}
\label{sec:edit}
After detecting bias attributes, our goal is to explicitly create bias-conflict samples, denoted as ${\Tilde{\vx}_{\text{bc}}}$, by editing a given sample in the biased dataset $\mathcal{D}_b$ to alter either its identified bias or target attribute. 
To convert bias-aligned samples into bias-conflict samples, we leverage a text-guided image editing models~\cite{liu2025step1x, zhang2024magicbrush, brooks2023instructpix2pix}, which allow image editing with simple natural language instructions.

To illustrate, in the BFFHQ dataset, where \{\texttt{class}, \texttt{bias}\} pairs are designed as \{\texttt{`age'}, \texttt{`gender'}\}, StaB detects \textit{\textbf{`woman'}} as the bias attribute for the \textit{\textbf{`young'}} class.
To edit such an image, we can construct an instruction like: \textit{``Turn this \textbf{young} person into a \textbf{young} \textbf{man} while keeping other attributes unchanged."}
This enables controllable editing of the bias attribute.

When producing bias-conflict samples, we not only edit bias attributes but also edit target attributes.
For example, changing a sample from the \textit{\textbf{`young'}} class with an instruction like \textit{``Turn this woman into an \textbf{old woman}.''} effectively creates bias-conflict samples of the \textit{\textbf{`old'}} class.
Notably, attributes such as \textbf{gender} (e.g., \textit{`woman'}) correspond to bias attributes automatically discovered by StaB.
While such target-edited samples do not constitute bias-conflict samples of the original \textit{\textbf{`young'}} class, they contribute to reducing the dependence between bias attributes and the class by creating bias-conflict sample of the other class (i.e., \textit{\textbf{`old'}}).

The process of image editing is clarified as follows:
\begin{gather}
\Tilde{\vx}_{\text{bc, be}}^{c}
= \text{ImageEditing}({\vx}^{c}, \mathcal{I}^{c}_{\text{be}}),
\label{eq:edit_be}
\\
\Tilde{\vx}_{\text{bc, te}}^{m}
= \text{ImageEditing}({\vx}^{c}, \mathcal{I}^{c\to m}_{\text{te}}).
\label{eq:edit_te}
\end{gather}

\begin{figure*}[ht!]
\centering
\includegraphics[width=\linewidth]{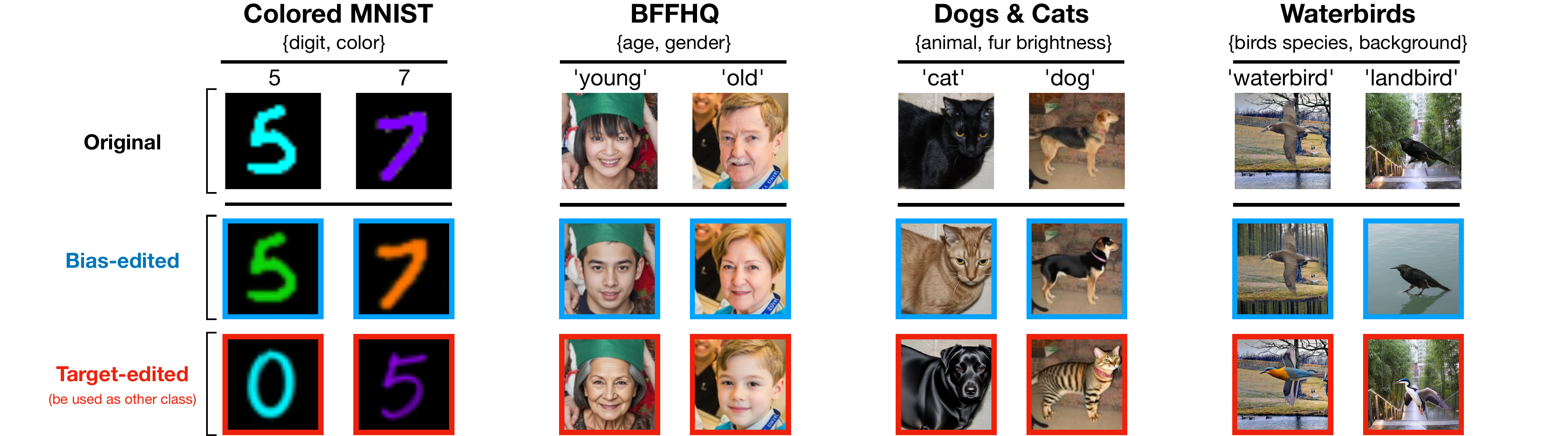}
\vspace{-0.5cm}
\caption{
\textbf{Bias-/target-attribute edited bias-conflict samples}.
Bias-conflict samples are produced by editing either the bias or target attributes from biased datasets.
Each dataset is denoted in the form \{class, bias\}.
Samples obtained by editing the bias attribute are indicated with a blue box, while those obtained by editing the target attribute are indicated with a red box.
}
\label{fig:edited_generated}
\vspace{-0.3cm}
\end{figure*}

$\mathcal{I}^{c}_{\text{be}}$ denotes an instruction for editing the bias attribute of a sample from class $c$; $\mathcal{I}^{c\to m}_{\text{te}}$ denotes an instruction for editing the target attribute of a sample from class $c$ to match the target of class $m$ ($m \neq c$).  
Accordingly, $\Tilde{\vx}_{\text{bc, be}}^{c}$ represents a bias-conflict sample for class $c$ produced by bias editing, while $\Tilde{\vx}_{\text{bc, te}}^{m}$ is a bias-conflict sample for class $m$ obtained by target editing a sample from class $c$.
The overall procedure for producing bias- and target-edited bias-conflict samples is illustrated in Figure~\ref{fig:main2}.

Notably, since we adopt an off-the-shelf pretrained text-guided editing model, neither additional model training nor supplementary training data is required during the construction of bias-conflict samples, making our approach efficient and widely applicable.

\subsection{A bias-reduced dataset construction}
\label{sec:bias-free-dataset}
The bias-reduced dataset $\hat{\mathcal{D}}$ is constructed by augmenting the given biased dataset $\mathcal{D}_b$ with bias-conflict samples produced through editing.
If only the edited bias-conflict samples are used, it is possible that new biases could be introduced or the original data distribution may be affected.
Thus, we combine the original dataset $\mathcal{D}_b$ with the created bias-conflict samples, resulting in a bias-mitigated dataset $\hat{\mathcal{D}}$.
The procedure for constructing the bias-mitigated dataset is summarized in Algorithm~\ref{alg:reconstructing_dataset}.

\begin{figure}[t]\centering
\vspace{-0.2cm}
{
\begin{minipage}[t]{1.0\columnwidth}\centering
\begin{algorithm}[H]
\caption{Constructing the bias-reduced dataset}
\label{alg:reconstructing_dataset}
\begin{algorithmic}[1]
\STATE \textbf{Input}: biased dataset $\mathcal{D}_b$, class set $\mathcal{C}$
\STATE \textbf{Output}: bias-reduced dataset $\hat{\mathcal{D}}$
\STATE Detect bias attributes:~Eqs.~\eqref{eq:extract-attrubute}--\eqref{eq:bias_attr}\\
    $\mathcal{B} = \{ a^c_{\text{bias}} \mid c \in \mathcal{C} \} \leftarrow \text{StaB}(\mathcal{D}_b)$
\STATE Construct editing instruction sets $\mathcal{I}_{\text{be}}$, $\mathcal{I}_{\text{te}}$ using $\mathcal{B}$
\STATE Initialize $\hat{\mathcal{D}} \gets \mathcal{D}_b$
\FOR{each sample ($\boldsymbol{x}_i, y_i) \in \mathcal{D}_b$}
    \STATE Produce bias-edited sample:~Eq.~\eqref{eq:edit_be}\\
      $\Tilde{\boldsymbol{x}}^{y_i}_{\text{bc, be}} \gets \text{ImageEditing}(\boldsymbol{x}^{y_i}_i, \mathcal{I}^{y_i}_{\text{be}})$
    \STATE Add $\Tilde{\boldsymbol{x}}^{y_i}_{\text{bc, be}}$ to $\hat{\mathcal{D}}$
    \STATE Sample $m \in \mathcal{C},\; m\neq y_i$
    \STATE Produce target-edited sample:~Eq.~\eqref{eq:edit_te}\\
      $\Tilde{\boldsymbol{x}}^{m}_{\text{bc, te}} \gets \text{ImageEditing}(\boldsymbol{x}^{y_i}_i, \mathcal{I}^{{y_i}\to m}_{\text{te}})$
    \STATE Add $\Tilde{\boldsymbol{x}}^{m}_{\text{bc, te}}$ to $\hat{\mathcal{D}}$
\ENDFOR
\STATE \textbf{Return} $\hat{\mathcal{D}}$
\end{algorithmic}
\end{algorithm}
\end{minipage}
}
\vspace{-0.4cm}
\end{figure}

\section{Experiments}
\subsection{Experimental settings}
\textbf{Datasets.}
In this work, to rigorously evaluate the effectiveness of BiasEdit, we conduct experiments on four widely used benchmark datasets: Colored MNIST~\cite{lee2021learning}, BFFHQ~\cite{kim2021biaswap}, Dogs \& Cats~\cite{kim2019learning} and Waterbirds~\cite{sagawa2019distributionally}.
Although these datasets are relatively small in scale, they serve as controlled and interpretable testbeds for studying dataset-induced bias—a fundamental issue that also arises in large Web-sourced data. 
Each dataset defines clear pairs of target and bias attributes (e.g., color–digit, gender–age, fur color-species, or background–species), allowing precise quantitative and qualitative assessment of how BiasEdit detects, edits, and mitigates biases under varying correlation strengths.
These benchmarks are widely adopted as standard proxies for bias analysis before methods are scaled to Web or industry-level data. 
Our focus is thus not on modeling specific Web datasets directly, but on validating a general and scalable mechanism that can operate without annotations or retraining. 
Because BiasEdit leverages pretrained vision–language and text-guided editing models, it can be seamlessly extended to larger, Web-sourced image corpora in future studies.

\noindent \textbf{Evaluation.}
Previous works~\cite{lee2023revisiting, lee2021learning, hwang2022selecmix} have chosen a final model based on the performance on the test set composed of only bias-conflict samples.

Furthermore, prior works have only reported the performance on bias-conflict samples in some datasets (namely, BFFHQ and Dogs \& Cats).
Such evaluation protocol leads to emphasis only on the performance on bias-conflict samples, not the overall generalization performance on both bias-conflict and bias-aligned samples.

On the other hand, we consider a practical evaluation protocol that favors a truly debiased model and reflects the process of model training and deployment in real-world settings.
Specifically, we choose the model based on the performance on validation drawn from the same distribution as the train set.
Then, the models are assessed on the both bias-aligned (BA) and bias-conflict (BC) samples.
Additionally, we evaluate using the same protocol as prior works~\cite{lee2023revisiting, lee2021learning, hwang2022selecmix} and report those results for comparison.

A lower ratio of bias-conflicting samples makes the debiasing task more challenging; key results focus on the most difficult setting (0\%) in the main paper, with additional results for 1\% and 5\% bias-conflict presented in the supplementary material.

\noindent \textbf{Baselines.}
We compare our method with eight diverse baselines that do not require bias annotations.  
The first baseline is a vanilla model~\cite{he2016deep} trained with cross-entropy loss.  
For reweighting methods, we consider LfF~\cite{nam2020learning} and its improved version BE~\cite{lee2023revisiting}, which reweight losses for bias-aligned and bias-conflict samples.  
We include SoftCon~\cite{hong2021unbiased} as a representative contrastive learning framework.  
Regarding augmentation approaches, DFA~\cite{lee2021learning} applies feature-level augmentation, while SelecMix~\cite{hwang2022selecmix} and BiasAdv~\cite{lim2023biasadv} employ image-level augmentation.  
Specifically, DFA swaps target and biased features; SelecMix uses mixup~\cite{zhang2017mixup}, and BiasAdv leverages adversarial attacks.  
Finally, DeNetDM~\cite{sreelatha2024denetdm} is a depth-modulation method designed to build robustness against spurious correlations.

\begin{table*}[t!]
\caption{\textbf{Image classification accuracy on bias-conflict (BC) and bias-aligned (BA) test sets, trained on a 100\% biased training set (0\% bias-conflict ratio).} 
For each dataset, BC denotes accuracy on bias-conflict samples, BA denotes accuracy on bias-aligned samples, and \textbf{Avg} denotes overall performance computed as the arithmetic mean of BC and BA. 
(a) reports the \emph{test set accuracy of the checkpoint with the best in-distribution validation performance}; (b) reports the \emph{test set accuracy of the checkpoint with the best bias-conflict test-set performance}. 
For each setting, we select the corresponding checkpoint as described above, compute group accuracies, and average results over 3 runs. The best results are in \textbf{bold}.}
\vspace{-0.2cm}
\centering

\begin{subtable}{\linewidth}
\centering
{\fontsize{9pt}{10pt}\selectfont
\renewcommand{\arraystretch}{1.1}
\setlength{\tabcolsep}{3pt}

\begin{tabularx}{\linewidth}{c | YYY | YYY | YYY | YYY}
\toprule
\multicolumn{13}{c}{Test set accuracy of the checkpoint with \textit{\textbf{the best in-distribution validation performance}}} \\
\midrule
Dataset & \multicolumn{3}{c}{Colored MNIST}
& \multicolumn{3}{c}{BFFHQ}
& \multicolumn{3}{c}{Dogs \& Cats}
& \multicolumn{3}{c}{Waterbirds} \\
\midrule
Group   & BC & BA & \cellcolor{gray!20}Avg
& BC & BA & \cellcolor{gray!20}Avg
& BC & BA & \cellcolor{gray!20}Avg
& BC & BA & \cellcolor{gray!20}Avg \\
\midrule
Vanilla~\cite{he2016deep}
& 0.56 & 99.96 & \cellcolor{gray!20}50.26 
& 37.96 & 99.32 & \cellcolor{gray!20}68.64 
& 20.03 & 97.32 & \cellcolor{gray!20}58.67
& 21.74 & 95.09 & \cellcolor{gray!20}58.42 \\
LfF~\cite{nam2020learning}
& 1.13 & 99.98 & \cellcolor{gray!20}50.55 
& 39.52 & 97.88 & \cellcolor{gray!20}68.70 
& 30.35 & 83.45 & \cellcolor{gray!20}56.90
& 39.91 & 83.47 & \cellcolor{gray!20}61.69 \\
LfF+BE~\cite{lee2023revisiting}
& 0.66 & 100.00 & \cellcolor{gray!20}50.33
& 34.20 & 90.53 & \cellcolor{gray!20}62.37
& 40.12 & 71.62 & \cellcolor{gray!20}55.87
& 63.11 & 48.37 & \cellcolor{gray!20}55.74 \\
SoftCon~\cite{hong2021unbiased}
& 0.12 & 100.00 & \cellcolor{gray!20}50.06
& 39.40 & 99.30 & \cellcolor{gray!20}69.35
& 23.63 & 97.81 & \cellcolor{gray!20}60.72
& 25.68 & 94.96 & \cellcolor{gray!20}60.32 \\
DeNetDM~\cite{sreelatha2024denetdm}
& 0.01 & 100.00 & \cellcolor{gray!20}50.00
& 49.13 & 99.07 & \cellcolor{gray!20}74.10
& 17.04 & 95.37 & \cellcolor{gray!20}56.21
& 21.70 & 94.78 & \cellcolor{gray!20}58.24 \\
DFA~\cite{lee2021learning}
& 0.81 & 100.00 & \cellcolor{gray!20}50.40
& 35.12 & 97.96 & \cellcolor{gray!20}66.54
& 23.81 & 91.20 & \cellcolor{gray!20}57.29
& 37.95 & 90.49 & \cellcolor{gray!20}64.22 \\
SelecMix~\cite{hwang2022selecmix}
& 2.22 & 99.93 & \cellcolor{gray!20}51.08
& 39.20 & 92.20 & \cellcolor{gray!20}65.70
& 32.38 & 72.30 & \cellcolor{gray!20}52.34
& 54.66 & 84.22 & \cellcolor{gray!20}69.44 \\
BiasAdv~\cite{lim2023biasadv}
& 2.88 & 99.97 & \cellcolor{gray!20}51.43
& 40.66 & 99.06 & \cellcolor{gray!20}69.86
& 15.41 & 95.12 & \cellcolor{gray!20}55.27
& 23.59 & 94.02 & \cellcolor{gray!20}58.81 \\
\midrule
\textbf{BiasEdit (ours)}
& 91.28 & 96.62 & \cellcolor{gray!20}\textbf{93.95}
& 62.46 & 98.93 & \cellcolor{gray!20}\textbf{80.70}
& 59.04 & 96.91 & \cellcolor{gray!20}\textbf{77.98}
& 58.28 & 91.49 & \cellcolor{gray!20}\textbf{74.88} \\
\bottomrule
\end{tabularx}
}
\caption{} 
\label{tab:main_table_a}
\end{subtable}

\begin{subtable}{\linewidth}
\centering
{\fontsize{9pt}{10pt}\selectfont
\renewcommand{\arraystretch}{1.1}
\setlength{\tabcolsep}{3pt}

\begin{tabularx}{\linewidth}{c | YYY | YYY | YYY | YYY}
\toprule
\multicolumn{13}{c}{Test set accuracy of the checkpoint with \textit{\textbf{the best bias-conflict test set performance}} (BA measured \emph{at the best BC})} \\
\midrule
Dataset & \multicolumn{3}{c}{Colored MNIST}
& \multicolumn{3}{c}{BFFHQ}
& \multicolumn{3}{c}{Dogs \& Cats}
& \multicolumn{3}{c}{Waterbirds} \\
\midrule
Group   & Best BC & BA & \cellcolor{gray!20}Avg
& Best BC & BA & \cellcolor{gray!20}Avg
& Best BC & BA & \cellcolor{gray!20}Avg
& Best BC & BA & \cellcolor{gray!20}Avg \\
\midrule
Vanilla~\cite{he2016deep}
& 0.98  & 99.96 & \cellcolor{gray!20}50.47
& 44.04 & 97.04 & \cellcolor{gray!20}70.54
& 25.83 & 93.51 & \cellcolor{gray!20}59.67
& 34.17 & 93.66 & \cellcolor{gray!20}63.92 \\
LfF~\cite{nam2020learning}
& 8.17  & 71.25 & \cellcolor{gray!20}39.71
& 46.80 & 67.88 & \cellcolor{gray!20}57.34
& 54.43 & 59.75 & \cellcolor{gray!20}57.09
& 71.58 & 63.64 & \cellcolor{gray!20}67.61 \\
LfF+BE~\cite{lee2023revisiting}
& 3.92  & 99.43 & \cellcolor{gray!20}51.68
& 49.53 & 67.33 & \cellcolor{gray!20}58.43
& 74.91 & 39.00 & \cellcolor{gray!20}56.96
& 75.02 & 38.18 & \cellcolor{gray!20}56.60 \\
SoftCon~\cite{hong2021unbiased}
& 1.35  & 99.93 & \cellcolor{gray!20}50.64
& 43.50 & 92.90 & \cellcolor{gray!20}68.20
& 33.06 & 85.68 & \cellcolor{gray!20}59.37
& 33.00 & 94.06 & \cellcolor{gray!20}63.53 \\
DeNetDM~\cite{sreelatha2024denetdm}
& 0.01  & 100.00 & \cellcolor{gray!20}50.00
& 62.80 & 98.26 & \cellcolor{gray!20}80.53
& 23.00 & 93.91 & \cellcolor{gray!20}58.46
& 33.25 & 93.40 & \cellcolor{gray!20}63.33 \\
DFA~\cite{lee2021learning}
& 4.62  & 99.80 & \cellcolor{gray!20}52.21
& 44.52 & 91.68 & \cellcolor{gray!20}68.10
& 45.95 & 73.36 & \cellcolor{gray!20}59.66
& 72.43 & 51.67 & \cellcolor{gray!20}62.05 \\
SelecMix~\cite{hwang2022selecmix}
& 9.76  & 78.36 & \cellcolor{gray!20}44.06
& 57.48 & 62.68 & \cellcolor{gray!20}60.08
& 80.98 & 29.03 & \cellcolor{gray!20}55.00
& 83.71 & 27.94 & \cellcolor{gray!20}55.83 \\
BiasAdv~\cite{lim2023biasadv}
& 4.25  & 100.00 & \cellcolor{gray!20}52.12
& 48.46 & 87.80 & \cellcolor{gray!20}68.13
& 25.91 & 88.33 & \cellcolor{gray!20}57.12
& 39.06 & 90.10 & \cellcolor{gray!20}64.58 \\
\midrule
\textbf{BiasEdit (ours)}
& 92.39 & 93.15 & \cellcolor{gray!20}\textbf{92.77}
& 71.33 & 97.33 & \cellcolor{gray!20}\textbf{84.33}
& 69.04 & 94.37 & \cellcolor{gray!20}\textbf{81.71}
& 69.03 & 83.07 & \cellcolor{gray!20}\textbf{76.05} \\
\bottomrule
\end{tabularx}%
}
\caption{} 
\label{tab:main_table_b}
\end{subtable}

\vspace{-0.5cm}
\label{tab:main_table}
\end{table*}

\noindent \textbf{Implementation details.}
All baseline experiments are conducted under identical experimental settings.
In particular, BiasEdit is trained with the same training setting as vanilla baseline, with only difference in the train set.
Thus, the performance difference introduced by BiasEdit arises solely from using the bias-reduced dataset $\hat{\mathcal{D}}$ as the train set.
The models used for each dataset are as follows: an three-layer MLP for Colored MNIST; a randomly initialized ResNet-18 for BFFHQ, Dogs \& Cats, and Waterbirds.
The in-distribution validation set is constructed by randomly sampling 20\% of the train set.
Learning rates are set to 0.01 for Colored MNIST, and 0.0001 for BFFHQ, Dogs \& Cats, and Waterbirds.
Images are resized to 28×28 pixels for Colored MNIST, while those from the other datasets are resized to 224×224 pixels.
No additional data augmentation is applied to Colored MNIST.
Random cropping and horizontal flipping are used as augmentation techniques for BFFHQ, Dogs \& Cats, and Waterbirds. 
Furthermore, all images are normalized per channel (3, H, W) with means of (0.4914, 0.4822, 0.4465) and standard deviations of (0.2023, 0.1994, 0.2010).
In BiasEdit, bias-attribute detection requires approximately 13 GB of GPU memory, and image editing brings the total up to 50 GB, so we run both on a single A100 GPU.
Once bias-conflict samples are created, the training of classifiers is performed on an RTX 2080Ti GPU.

\subsection{Main results}
\label{sec:main_results}
We evaluate each method on bias-aligned and bias-conflict test sets under the most challenging setting (0\% bias-conflict ratio; i.e., 100\% biased) and report the results in Table~\ref{tab:main_table}.
Evaluation across various bias-conflict ratios and the results obtained using the evaluation protocols of previous studies~\cite{kim2022learning, hwang2022selecmix} are reported in the appendix. 

\noindent \textbf{Analysis on 100\% biased datasets.}
In a fully biased train set (i.e., with a 0\% bias-conflict ratio), where bias and target attributes are completely spuriously correlated, distinguishing between bias and target attributes becomes inherently challenging.
Consequently, baseline methods perform very poorly, often exhibiting accuracy on bias-conflict samples that falls below random guessing.
Meanwhile, baseline methods exhibit very high performance on bias-aligned samples (BA), demonstrating that baseline methods fail to learn a debiased classifier with a fully biased train set.

\begin{figure}[t!]
\centering
\includegraphics[width=\linewidth]{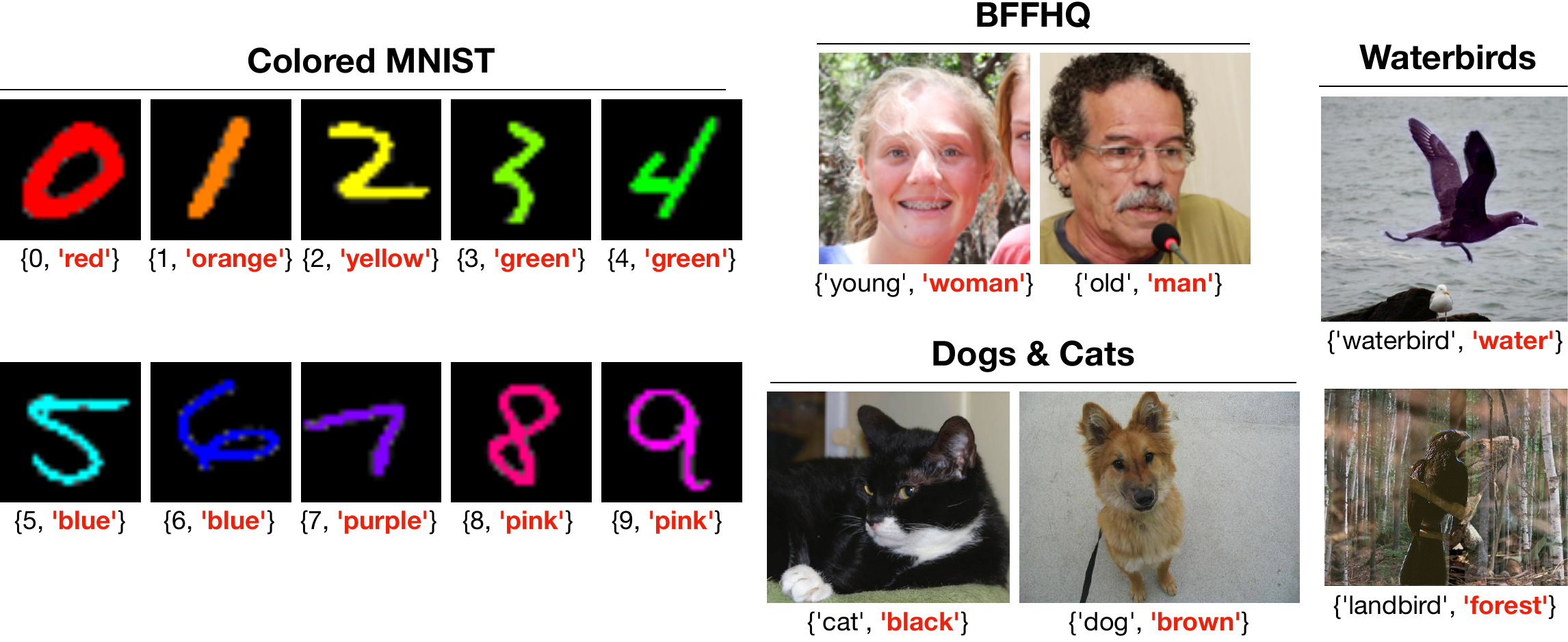}
\caption{\textbf{Bias attributes detected by StaB}. For four datasets, the detected class and bias attributes (in red) are shown.}
\label{fig:tags}
\end{figure}

By contrast, BiasEdit significantly improves the performance on bias-conflict samples and the overall performance, despite the lack of bias-conflict samples in the original train set.
The results demonstrate the effectiveness of our proposed framework that detects and edits attributes to create diverse bias-conflict samples, particularly in scenarios where a training dataset is highly biased.
Figure~\ref{fig:edited_generated} shows bias-conflict samples edited for bias and target attributes

Furthermore, to select a model with higher average performance, we note that a model should be selected based on the performance on bias-reduced validation set, not a biased one.  
Since BiasEdit allows constructing a bias-reduced dataset, it enables creating a suitable validation set for unknown-bias debiasing, facilitating robust model selection. We also evaluate under the prior protocol of selecting the checkpoint with the best bias-conflict test set accuracy, and find that the overall performance of BiasEdit remains superior to that of baselines.

\noindent \textbf{Bias attributes detected by StaB module.}
We report the bias attributes detected by the StaB module for each benchmark dataset in Figure~\ref{fig:tags}.
We analyze that our proposed StaB module is capable of detecting interpretable bias attributes not only in synthetic datasets such as Colored MNIST and Waterbirds, but also in datasets composed of natural images such as BFFHQ and Dogs \& Cats.
In the case of Colored MNIST, although the subtle color differences between classes are not easily described in text, they serve as appropriate bias attributes. 
For Dogs \& Cats, the detected bias attribute captures information related to fur brightness.
We report the results of bias attribute detection for train sets with various bias-conflict ratios in the appendix section.

\subsection{Ablation studies}
\noindent \textbf{StaB module conditions and bias attributes.}
This ablation study examines the effects of statistical dependence and the mutual information condition in the StaB module.
Table~\ref{tab:ablation_tags} presents the bias attributes detected under different combinations of these conditions.
When only the statistical dependence condition is used, rare attributes tend to be selected as biases.
In contrast, applying only the mutual information condition does not consider entropy reduction at the class level.
Appropriate bias attributes are selected only when both conditions are applied together.

\begin{table}[t!]
\caption{
\textbf{Ablation study on StaB conditions} with 0\% bias-conflict ratio datasets.
\(\mathbf{P}_{\mathrm{dp}}\) indicates whether the statistical dependence condition (Condition 1) is met.
\(\text{MI}\) denotes the use of mutual information as a bias detection criterion (Condition 2).
\boldcheckmark indicates presence; - indicates absence.
}
\label{tab:ablation_tags}
\small
\centering
\setlength{\tabcolsep}{0.3em}
\renewcommand{\arraystretch}{1.2}

\begin{tabular}{cc|cc|cc|cc}
    \toprule
    \multirow{2}{*}{\textbf{$\mathbf{P}_{\mathrm{dp}}$}} &
    \multirow{2}{*}{\textbf{$\text{MI}$}} &
    \multicolumn{2}{c|}{CMNIST} &
    \multicolumn{2}{c|}{BFFHQ} &
    \multicolumn{2}{c}{Waterbirds} \\
    \cmidrule{3-4} \cmidrule{5-6} \cmidrule{7-8}
    & & 0 & 1 & `young' & `old' & `waterbird' & `landbird' \\
    \midrule
    \boldcheckmark & -              & `apple' & `stick' & `snacks' & `brazil' & `smooth' & `wavy' \\
    -              & \boldcheckmark & `pink'  & `pink'  & `man'    & `man'    & `water'  & `water' \\
    \boldcheckmark & \boldcheckmark & \textbf{`red'} & \textbf{`orange'} & \textbf{`woman'} & \textbf{`man'} & \textbf{`water'} & \textbf{`forest'} \\
    \bottomrule
\end{tabular}
\end{table}
\begin{table}[t!]
    \vspace{0.5cm}
    \caption{
        \textbf{Group average accuracy on the bias-reduced train set $\hat{\mathcal{D}}_{\text{sampled}}$ of the same size as the biased dataset $\mathcal{D}_b$.}
        BiasEdit provides large performance improvement even when the bias-reduced train set is the same size as the biased dataset.
    }
    \label{tab:sampled_datasets}
    \centering

    {\renewcommand{\arraystretch}{1.2}
    \setlength{\tabcolsep}{4pt}
    \begin{tabularx}{\linewidth}{>{\centering\arraybackslash}X|YYYY}
        \toprule
        Dataset & CMNIST & BFFHQ & D\&C & Waterbirds \\
        \midrule
        $\mathcal{D}_b$ & 50.26\% & 68.64\% & 58.67\% & 58.42\% \\
        \midrule
        $\hat{\mathcal{D}}_{\text{sampled}}$ & 92.70\% & 80.33\% & 73.10\% & 74.07\% \\
        $\hat{\mathcal{D}}$ (ours) & \textbf{93.95\%} & \textbf{80.70\%} & \textbf{77.98\%} & \textbf{74.88\%} \\
        \bottomrule
    \end{tabularx}
    }
\end{table}

\noindent \textbf{Comparison of performance according to dataset size.}
The bias-reduced dataset $\hat{\mathcal{D}}$ includes additional bias-conflict samples created by BiasEdit.
One may argue that the performance improvement comes from additional data.
To assess the influence of dataset size on the performance, we evaluate the performance by constructing a bias-reduced dataset $\hat{\mathcal{D}}_{\text{sampled}}$ that matches the original dataset in size in Table~\ref{tab:sampled_datasets}.
Specifically, after constructing the bias-reduced dataset $\hat{\mathcal{D}}$, we randomly sample from it to obtain a train set $\hat{\mathcal{D}}_{\text{sampled}}$ of the same size as the original dataset ${\mathcal{D}}_b$.
Although $\hat{\mathcal{D}}_{\text{sampled}}$ created by BiasEdit is matched in size to the original dataset, it still demonstrates effective debiasing performance and outperforms other baseline methods in most cases.
Furthermore, the group average accuracy of $\hat{\mathcal{D}}_{\text{sampled}}$ shows only minimal decreases in most cases related to $\hat{\mathcal{D}}$.
The results demonstrate that the most of performance improvement brought by BiasEdit does not come from additional data, but rather created bias-conflict samples themselves, further corroborating the effectiveness of the process of detecting and editing bias attributes in BiasEdit.

\begin{table}[t!]
\caption{\textbf{Analysis of group average accuracy according to bias-/target-edited samples} on datasets with a 0\% bias-conflict ratio. Both bias-edited and target-edited samples demonstrate effective results for debiasing. A checkmark (\cmark) indicates that the corresponding sample type is included in the train set, while a dash (-) denotes its exclusion.}
\centering
\def\arraystretch{1.2}
\begin{tabularx}{\columnwidth}{c c c |
>{\centering\arraybackslash}X
>{\centering\arraybackslash}X
>{\centering\arraybackslash}X
>{\centering\arraybackslash}X }
\toprule
$\mathcal{D}_b$  & $\hat{\mathcal{D}}_{\text{bc,be}}$  & $\hat{\mathcal{D}}_{\text{bc,te}}$  
    & {\footnotesize CMNIST}  & {\footnotesize BFFHQ}  & {\footnotesize D\&C}  & {\footnotesize Waterbirds} \\
\midrule
\cmark  & -      & -      & 50.26\%  & 68.64\%  & 58.67\%  & 58.42\% \\
\midrule
\cmark  & \cmark & -      & \textbf{94.40}\%  & 78.93\%  & 73.38\%  & 66.87\% \\
\cmark  & -      & \cmark & 63.10\%  & 75.46\%  & 68.97\%  & 72.02\% \\
\midrule
\cmark  & \cmark & \cmark & 93.95\%  & \textbf{80.70}\%  & \textbf{77.98}\%  & \textbf{74.88}\% \\
\bottomrule
\end{tabularx}
\label{tab:bias_target_edited}
\end{table}

\noindent \textbf{Analysis of the debiasing effects of bias-edited samples and target-edited samples.}
Table~\ref{tab:bias_target_edited} showcases the effectiveness of the bias-edited samples, $\hat{\mathcal{D}}_{\text{bc, be}}$, and the target-edited samples, $\hat{\mathcal{D}}_{\text{bc, te}}$.
Consistently, both $\hat{\mathcal{D}}_{\text{bc, be}}$ and $\hat{\mathcal{D}}_{\text{bc, te}}$, when applied independently, lead to the performance improvement over ${\mathcal{D}}_b$.
Moreover, applying both $\hat{\mathcal{D}}_{\text{bc, be}}$ and $\hat{\mathcal{D}}_{\text{bc, te}}$ together yields the highest performance in most cases.

\begin{table}[t!]
\caption{
\textbf{Comparison of debiasing performance using bias-conflict samples generated by generative models and text-guided editing models.}
Dataset $\mathcal{\hat{D}}_g$, constructed with bias-conflict samples generated via a T2I generative model, shows little or no improvement in performance.
By contrast, our dataset $\hat{\mathcal{D}}$, composed of bias-conflict samples produced by editing, yields substantial performance improvement.
}
\vspace{-0.2cm}
\centering
\def\arraystretch{1.2}
\begin{tabularx}{\columnwidth}{c |
>{\centering\arraybackslash}X
>{\centering\arraybackslash}X
>{\centering\arraybackslash}X
>{\centering\arraybackslash}X }
\toprule
Dataset & {\footnotesize CMNIST} & {\footnotesize BFFHQ} & {\footnotesize D\&C} & {\footnotesize Waterbirds} \\
\midrule
$\mathcal{D}_b$    & 50.26\% & 68.64\% & 58.67\% & 58.42\% \\
$\mathcal{\hat{D}}_g$    & 69.05\% & 62.02\% & 61.79\% & 62.02\% \\
\midrule
$\hat{\mathcal{D}}$ (ours) & \textbf{93.95\%} & \textbf{80.70\%} & \textbf{77.98\%} & \textbf{74.88\%} \\
\bottomrule
\end{tabularx}
\label{tab:generated}
    \vspace{-0.2cm}
\end{table}

\noindent \textbf{Why \textit{edit}? Why not \textit{generate}?}
To answer the question and justify our design choice, we compare the performance when images are either edited or generated to produce bias-conflict samples.
To this end, we generate bias-conflict samples with a Text-to-Image (T2I) generative model~\cite{esser2024scaling}.
To maximize the preservation of image information, bias-/target-edited captions are generated with a Image-to-Text model~\cite{liu2024improved} and used as input to the T2I generative model.
Table~\ref{tab:generated} reports the performance of a classifier trained on a new dataset $\mathcal{\hat{D}}_g$, obtained by augmenting a train set with the generated bias-conflict samples.

\begin{figure}[t!]
    \centering
    \includegraphics[width=\linewidth]{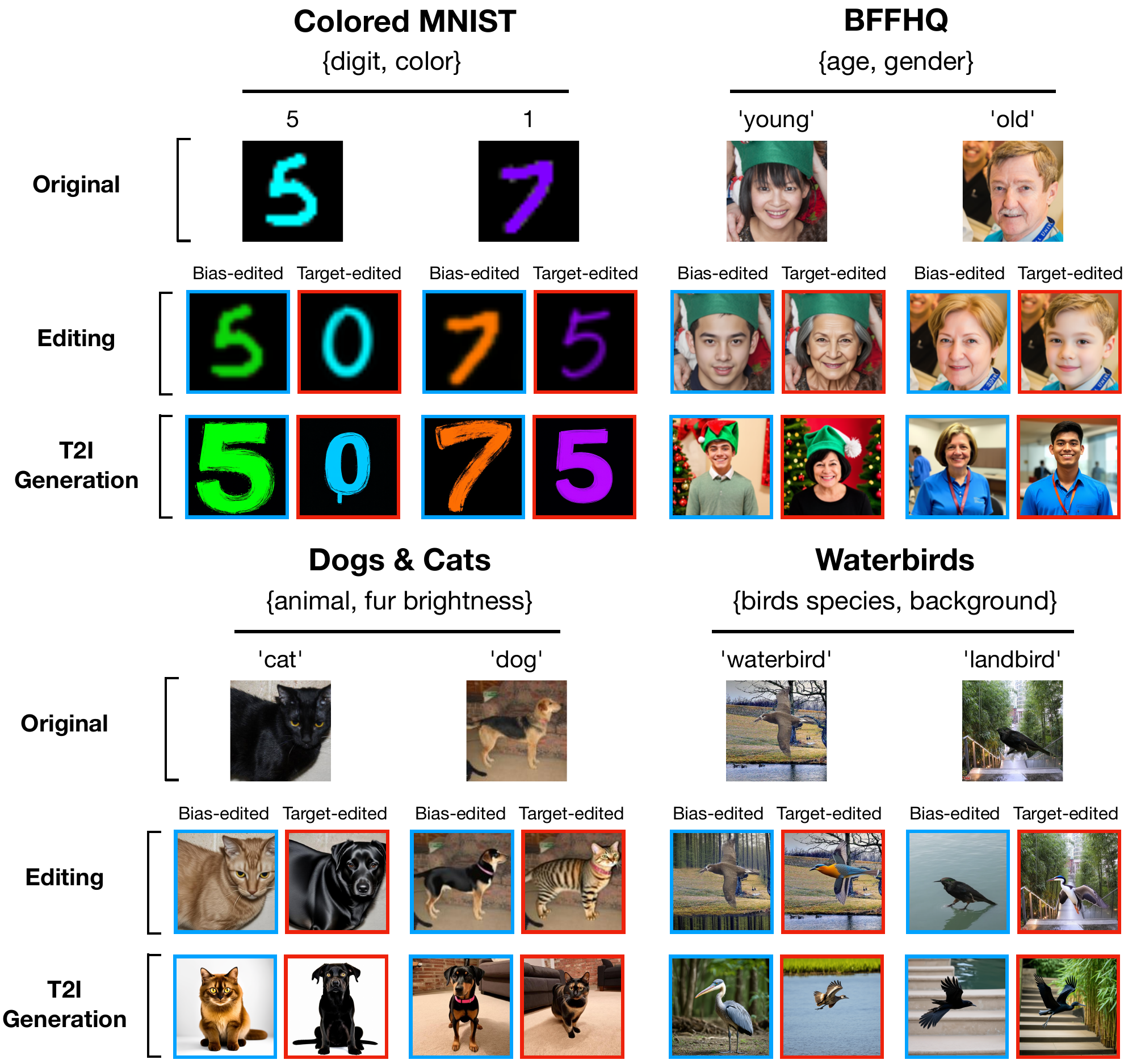}
    \vspace{-0.6cm}
    \caption{
    \textbf{Illustration of the visual comparison between bias-conflict samples created using an image editing model and those generated by a generation model.}
    Bias-conflict samples created with the image editing model tend to preserve various attributes of the original image while primarily editing the bias or target attribute.
    In contrast, samples generated by the generation model fail to retain the attributes of the original image as well. 
    Bias-edited samples are indicated with blue boxes, while target-edited samples with red boxes.
    }
    \label{fig:generation_compare}
\end{figure}

The results demonstrate that editing attributes in images lead to substantial performance improvement, compared to directly generating bias-conflict samples.
Given textual descriptions, an image generator struggles to capture the subtle data distribution of the original dataset, especially nuanced attributes that are difficult to express in language.
By contrast, the editing approach can be more effective at maintaining the distribution, as it focuses on editing the necessary attributes while largely preserving other aspects of the image.

In Figure~\ref{fig:generation_compare}, we visually compare bias-conflict samples created using an image editing model~\cite{liu2025step1x} with those generated by a Text-to-Image (T2I) model~\cite{esser2024scaling}.
Samples created with the image editing model tend to preserve much of the original image information while modifying the necessary attributes (i.e., bias and target attributes).
In contrast, samples generated by the T2I model often fail to retain the details of the original image.
Notably, T2I-generated samples frequently differ from the original in aspects that are difficult to describe linguistically, such as pose or camera angle.
Therefore, bias-conflict samples created via image editing models are advantageous for maintaining the distribution of the original dataset compared to those generated by T2I models.

\noindent\textbf{Multi-bias.} We demonstrate that BiasEdit can handle co-occurring biases using a multi-bias CelebA~\cite{liu2015deep} setup with gender and age attributes, where it jointly mitigates multiple bias factors via text-guided editing; further details are provided in Appendix~\ref{multi_bias}.

\section{Conclusion}
We presented BiasEdit, a modular and training-free framework for learning debiased classifiers through bias attribute detection and editing. 
By combining statistical analysis of visual–linguistic attributes with text-guided image editing, BiasEdit generates realistic bias-conflict samples that effectively mitigate spurious correlations in biased datasets. 
Our experiments demonstrate that BiasEdit substantially improves generalization and fairness even in extreme conditions—such as when all training samples are bias-aligned—outperforming existing debiasing methods without requiring manual annotation or additional model training.
Although our evaluation focused on benchmark datasets, the framework is designed to generalize to large, Web-sourced image collections. 
By operating on top of existing vision–language and image editing models, BiasEdit provides a practical step toward trustworthy, fair, and robust visual AI systems used in Web services such as content moderation, recommendation, and retrieval. 
We view BiasEdit as a promising building block for responsible Web intelligence, where models trained on Web data not only achieve high accuracy but also uphold principles of fairness and accountability.
\section{Acknowledgments}
This work was supported by the Institute of Information \& Communications Technology Planning \& Evaluation (IITP) grant funded by the Korea government (MSIT) (No. RS-2020-II201373, Artificial Intelligence Graduate School Program (Department of Artificial Intelligence, Hanyang University)) and by the National Research Foundation of Korea (NRF) grant funded by the Korea government (MSIT) (No. RS-2025-24533064). 

\clearpage

\bibliographystyle{ACM-Reference-Format}
\bibliography{sample-base}
\clearpage

\appendix
\FloatBarrier

\section{Bias attribute detection by StaB}
In this section, we present the bias attributes detected by StaB at bias-conflict ratios of 0\%, 1\%, and 5\% in Table~\ref{tab:stab_bias}.
Additionally, we show the top-3 attributes ranked by mutual information in descending order.
StaB robustly detects bias attributes with minimal variation in results despite changes in the bias-conflict ratio.
An interesting observation is that, in the BFFHQ dataset, the attributes following the top-ranked mutual information attribute commonly correlate with gender-related keywords (e.g., `dress', `tie').
In the Waterbirds dataset, the attributes with high mutual information all correspond to background-related features.
\FloatBarrier
\begin{table}[ht!]
    \centering
    \caption{
        \textbf{StaB‐detected bias attributes across four datasets.}
        For each bias-conflict ratio (0\%, 1\%, 5\%), the top-3 attributes ranked by mutual information are shown for each dataset. StaB consistently identifies appropriate bias attributes across diverse datasets and bias-conflict conditions.
    }
    \begin{subtable}[t]{0.48\textwidth}
        \renewcommand{\arraystretch}{1.1}
        \centering
        \scalebox{1}{
            \begin{tabular}{c|c|c}
                \toprule
                \makecell{Bias-conflict\\ratio}
                    & \makecell{\textbf{Colored MNIST}\\(0 / 1)}
                    & \makecell{\textbf{BFFHQ}\\(`young' / `old')} \\
                \midrule
                \multirow{3}{*}{0.0\%}
                    & \textbf{`red' / `orange'}          & \textbf{`woman' / `man'}    \\
                    & `black' / `black'                 & `hair' / `suit'            \\
                    & `pixelated' / `thin'              & `dress' / `tie'            \\
                \midrule
                \multirow{3}{*}{1.0\%}
                    & \textbf{`red' / `orange'}          & \textbf{`woman' / `man'}    \\
                    & `black' / `black'                 & `hair' / `suit'            \\
                    & `circle' / `thin'                 & `dress' / `tie'            \\
                \midrule
                \multirow{3}{*}{5.0\%}
                    & \textbf{`red' / `orange'}          & \textbf{`woman' / `man'}    \\
                    & `black' / `black'                 & `hair' / `suit'            \\
                    & `circle' / `thin'                 & `dress' / `tie'            \\
                \bottomrule
            \end{tabular}
        }
        \caption{StaB‐detected bias attributes on Colored MNIST and BFFHQ.}
        \label{tab:stab_bias_attrs_part1}
    \end{subtable}
    \hfill
    \begin{subtable}[t]{0.48\textwidth}
        \renewcommand{\arraystretch}{1.1}
        \centering
        \scalebox{1}{
            \begin{tabular}{c|c|c}
                \toprule
                \makecell{Bias-conflict\\ratio}
                    & \makecell{\textbf{Dogs \& Cats}\\(`cat' / `dog')}
                    & \makecell{\textbf{Waterbirds}\\(`landbird' / `waterbird')} \\
                \midrule
                \multirow{3}{*}{0.0\%}
                    & \textbf{`black' / `brown'}         & \textbf{`forest' / `water'}   \\
                    & `gray' / `leash'                  & `leaves' / `ocean'           \\
                    & `sit' / `grass'                   & `green' / `sea'              \\
                \midrule
                \multirow{3}{*}{1.0\%}
                    & \textbf{`black' / `brown'}         & \textbf{`forest' / `water'}   \\
                    & `gray' / `leash'                  & `leaves' / `ocean'           \\
                    & `sit' / `grass'                   & `green' / `sea'              \\
                \midrule
                \multirow{3}{*}{5.0\%}
                    & \textbf{`black' / `brown'}         & \textbf{`forest' / `water'}   \\
                    & `gray' / `leash'                  & `green' / `ocean'            \\
                    & `sit' / `grass'                   & `leaves' / `sea'             \\
                \bottomrule
            \end{tabular}
        }
        \caption{StaB‐detected bias attributes on Dogs \& Cats and Waterbirds.}
        \label{tab:stab_bias_attrs_part2}
    \end{subtable}
    \label{tab:stab_bias}
\end{table}

\section{Debiasing multiple biases}
\label{multi_bias}

In this section, we demonstrate the ability of BiasEdit to mitigate spurious correlations when multiple biases coexist in the training data. Following the evaluation protocol of prior work~\cite{kappiyath2025sebra}, we evaluate BiasEdit on a reconstructed version of the CelebA~\cite{liu2015deep} dataset in which `smiling' (vs. `not smiling') is the target attribute and the datasets have been deliberately reconstructed to exhibit biases with respect to both gender and age.

Concretely, the restructured CelebA setup exhibits a strong correlation between `smiling' and the attributes `female' and `young', while `not smiling' is correlated with `male' and `not young'.
Using the StaB bias-discovery method, the top two bias attributes detected for each class are confirmed to be \{`smiling': `female', `young'\} and \{`not smiling': `male', `middle-aged'\}.

Table~\ref{tab:celeba} compares the performance of the baseline and BiasEdit on the CelebA dataset containing multiple biases. 
Even in environments with multiple biases, StaB effectively captures and mitigates various forms of bias simultaneously, demonstrating superior performance compared to the baseline.

\FloatBarrier
\begin{table}[t!]
\caption{Comparison of debiasing performance on the CelebA dataset containing multiple biases. ID Acc denotes the accuracy on a test set that follows the same distribution as the training set, while WG Acc represents the accuracy of the worst-performing group.}
\vspace{-0.2cm}
\centering
\def\arraystretch{1.0}
\begin{tabularx}{\columnwidth}{
  >{\centering\arraybackslash}X
  | >{\centering\arraybackslash}X
  >{\columncolor{gray!20}\centering\arraybackslash}X
}
\toprule
Method & ID Acc & WG Acc \\
\midrule
Vanilla~\cite{he2016deep}  & 96.62 & 46.90  \\
LfF~\cite{nam2020learning}  & 62.15 & 51.08  \\
LfF+BE~\cite{lee2023revisiting}  & 38.17 & 34.83  \\
SoftConn~\cite{hong2021unbiased}  & 96.89 & 51.75  \\
DeNetDM~\cite{sreelatha2024denetdm}  & 96.17 & 47.58  \\
DFA~\cite{lee2021learning}  & 79.56 & 53.75  \\
SelecMix~\cite{hwang2022selecmix}  & 79.81 & 50.41  \\
BiasAdv~\cite{lim2023biasadv}  & 95.58 & 46.25 \\
\midrule
BiasEdit (ours)  & 96.53 & \textbf{63.41}  \\
\bottomrule
\end{tabularx}
\label{tab:celeba}
\vspace{-0.2cm}
\end{table}

\section{Bias-discovery comparison}
In this section, we compare the proposed StaB module with B2T~\cite{kim2024discovering} from the perspective of bias discovery in Table~\ref{tab:b2t_comp}.
While B2T detects bias based on samples where the model fails to make correct predictions, it does not directly identify the bias itself and shows large variations in performance depending on the bias-conflict ratio.
In contrast, StaB discovers biases directly from the dataset, thereby identifying the biases themselves, maintaining consistency across different bias-conflict ratios, and producing more valid and reliable results than B2T.

\FloatBarrier
\begin{table}[ht!]
    \centering
    \caption{
        \textbf{Comparison of bias discovery between B2T and StaB.}
        Compared to B2T, StaB not only achieves superior bias discovery performance but also maintains consistency across different bias-conflict ratios.
    }
    \begin{subtable}[t]{0.48\textwidth}
        \renewcommand{\arraystretch}{1.1}
        \centering
        \scalebox{1}{
            \begin{tabular}{c|c|c}
                \toprule
                \makecell{Bias-conflict\\ratio}
                    & \makecell{\textbf{B2T} \\ (`young', `old')}
                    & \makecell{\textbf{StaB (ours)} \\ (`young', `old')} \\
                \midrule
                \multirow{3}{*}{0.0\%}
                    & \textbf{`glasses' / `jugglings'}          & \textbf{`woman' / `man'}    \\
                    & `long day' / `flower'                 & `hair' / `suit'            \\
                    & `day of work' / `long beard'              & `dress' / `tie'            \\
                \midrule
                \multirow{3}{*}{1.0\%}
                    & \textbf{`man' / `hat'}          & \textbf{`woman' / `man'}    \\
                    & `camera' / `teacher'                 & `hair' / `suit'            \\
                    & `glasses' / `clown'                 & `dress' / `tie'            \\
                \midrule
                \multirow{3}{*}{5.0\%}
                    & \textbf{`man' / `woman'}          & \textbf{`woman' / `man'}    \\
                    & `beard' / `graduate'                 & `hair' / `suit'            \\
                    & `long' / `black'                 & `dress' / `tie'            \\
                \bottomrule
            \end{tabular}
        }
        \caption{Comparison of bias attributes detected by B2T and StaB (ours) on BFFHQ under various bias-conflict ratios}
        \label{tab:stab_bias_attrs_part1}
    \end{subtable}
    \hfill
    \begin{subtable}[t]{0.48\textwidth}
        \renewcommand{\arraystretch}{1.1}
        \centering
        \scalebox{1}{
            \begin{tabular}{c|c|c}
                \toprule
                \makecell{Bias-conflict\\ratio}
                    & \makecell{\textbf{B2T} \\ (`landbird', `waterbird')}
                    & \makecell{\textbf{StaB (ours)} \\ (`landbird', `waterbird')} \\
                \midrule
                \multirow{3}{*}{0.0\%}
                    & \textbf{`flies' / `pond'}          & \textbf{`forest' / `water'}    \\
                    & `city' / `woods'                 & `leaves' / `ocean'            \\
                    & `park' / `tree branch'              & `green' / `sea'            \\
                \midrule
                \multirow{3}{*}{1.0\%}
                    & \textbf{`water' / `woods'}          & \textbf{`forest' / `water'}    \\
                    & `person' / `pond'                 & `leaves' / `ocean'            \\
                    & `flying' / `field'                 & `green' / `sea'            \\
                \midrule
                \multirow{3}{*}{5.0\%}
                    & \textbf{`beach' / `pond'}          & \textbf{`forest' / `water'}    \\
                    & `water' / `tree'                 & `green' / `ocean'            \\
                    & `dock' / `branch'                 & `leaves' / `sea'            \\
                \bottomrule
            \end{tabular}
        }
        \caption{Comparison of bias attributes detected by B2T and StaB (ours) on Waterbirds under various bias-conflict ratios}
        \label{tab:stab_bias_attrs_part2}
    \end{subtable}

    \label{tab:b2t_comp}
\end{table}

\section{Flexibility of BiasEdit}
We demonstrate that BiasEdit is BiasEdit is a flexible, modular framework for leveraging off-the-shelf vision–language and image editing models.
Table~\ref{tab:mb_results} compares the dataset $\mathcal{\hat{D}}_{\text{IP2P}}$, constructed by editing biases using the vision–language model Tag2Text~\cite{huang2023tag2text} and the editing model Instruct-Pix2Pix~\cite{brooks2023instructpix2pix, zhang2024magicbrush}, with $\mathcal{\hat{D}}_{\text{Step1X}}$, the bias-edited dataset used in the main results leveraging LLaVA~\cite{liu2024improved} and Step-1X~\cite{liu2025step1x}; both datasets are evaluated on the BFFHQ dataset at a 0\% bias-conflict ratio.
As more advanced vision–language and editing models are employed, consistent improvements in performance are observed, suggesting that the debiasing capability of BiasEdit can further benefit from future advancements in pretrained model performance.

\begin{table}[ht!]
    \centering
    \caption{Comparison of debiasing performance using different vision--language and image-editing models. BiasEdit leverages off-the-shelf pretrained models, making it a flexible framework, and debiasing performance improves as stronger pretrained models are used.}
    \label{tab:mb_results}
    \renewcommand{\arraystretch}{1.1}

    \begin{tabular}{>{\centering\arraybackslash}m{1cm}|cc>{\columncolor{gray!20}}c}
        \toprule
        Dataset & BC & BA & Avg \\
        \midrule
        $\mathcal{D}_b$ & 37.96\% & 99.32\% & 68.64\% \\
        \midrule
        $\hat{\mathcal{D}}_{\text{IP2P}}$ & 45.40\% & 99.60\% & 72.50\% \\
        $\hat{\mathcal{D}}_{\text{Step1X}}$ & 58.53\% & 99.33\% & 78.93\% \\
        \bottomrule
    \end{tabular}
\end{table}

\section{Scalability and perceptual quality}
To assess scalability beyond curated benchmarks, we evaluate BiasEdit on a LAION subset (LAION-occupation)~\cite{friedrich2023fair} for a doctor--nurse classification task under the unknown-bias setting. StaB automatically discovers a `male' bias for the `doctor' class and a `female' bias for the `nurse' class, and BiasEdit mitigates these biases via text-guided editing.

\begin{table}[t!]
    \centering
    \caption{Scalability and perceptual quality on LAION-occupation.}
    \label{tab:laion}

    \begin{subtable}[t]{\linewidth}
        \centering
        \caption{Comparison of debiasing performance on LAION-occupation.}
        \label{tab:laion_scalability}
        \renewcommand{\arraystretch}{1.1}
        \begin{tabular}{>{\centering\arraybackslash}m{2cm}|cc>{\columncolor{gray!20}}c}
            \toprule
            Method & BC & BA & Avg \\
            \midrule
            Vanilla~\cite{he2016deep} & 43.98\% & 92.42\% & 68.20\% \\
            DeNetDM~\cite{sreelatha2024denetdm} & 49.32\% & 89.46\% & 69.39\% \\
            \midrule
            BiasEdit (ours) & 53.08\% & 88.34\% & \textbf{70.71\%} \\
            \bottomrule
        \end{tabular}
    \end{subtable}

    \medskip

    \begin{subtable}[t]{\linewidth}
        \centering
        \caption{User study scores on LAION-occupation (5-point scale).}
        \label{tab:laion_perceptual}
        \renewcommand{\arraystretch}{1.1}
        \begin{tabular}{ccc>{\columncolor{gray!20}}c}
            \toprule
            Correctness & Preservation & Naturalness & Avg \\
            \midrule
            4.116 & 4.345 & 4.326 & 4.262 \\
            \bottomrule
        \end{tabular}
    \end{subtable}
\end{table}

Table~\ref{tab:laion_scalability} compares BC/BA average-accuracy against the vanilla baseline and the SOTA baseline DeNetDM~\cite{sreelatha2024denetdm}, demonstrating effective debiasing on noisy web data.

To evaluate perceptual quality, we conduct a user study on 80 randomly sampled edited LAION images without manual selection, rated by 22 participants. Participants assess correctness of the intended attribute change, preservation of other attributes, and visual naturalness; Table~\ref{tab:laion_perceptual} summarizes the results.

\section{Additional results}
In this section, we present additional results to supplement the main paper.
Tables~\ref{tab:additional_results} provide debiasing performance evaluations for the Colored MNIST~\cite{lee2021learning}, BFFHQ~\cite{kim2021biaswap}, Dogs \& Cats~\cite{kim2019learning}, and Waterbirds~\cite{sagawa2019distributionally} datasets across various bias-conflict ratios. 
Table~\ref{tab:additional_results} reports results at the checkpoint that achieves the best in-distribution validation performance, and — following the evaluation protocol of prior works~\cite{hwang2022selecmix,lee2023revisiting} — also reports results at the checkpoint that yields the highest bias-conflict test set accuracy.
In both tables we present the test accuracies for bias-aligned (BA), bias-conflicting (BC), and the group average (Avg).
In addition, we report the highest observed bias-conflicting accuracy (Best BC) and the corresponding bias-aligned accuracy and group average measured at that Best BC.
This protocol appropriately evaluates overall group generalization performance without overemphasizing the bias-conflict group.

Notably, BiasEdit achieves superior group average accuracy compared to all baselines across all bias-conflict ratios (0\%, 1\%, and 5\%).
On the BFFHQ, Dogs \& Cats and Waterbirds datasets, SelecMix~\cite{hwang2022selecmix} achieves the highest bias-conflict test set accuracy (Best BC) but incurs a substantial drop in bias-aligned performance, revealing an overemphasis on the bias-conflict group.
This evaluation protocol—whether selecting by Best BC or by in-distribution validation performance—fails to capture overall group generalization, as evidenced by degraded bias-aligned accuracy on BFFHQ, Dogs \& Cats, and Waterbirds.
In contrast, BiasEdit consistently outperforms all baselines across all four benchmark datasets under both realistic model selection criteria, demonstrating robust generalization and practical debiasing effectiveness.

\FloatBarrier
\begin{table*}[h]
\caption{Performance evaluation across different bias-conflict ratios. For bias-conflict ratios of 1\% and 5\%, we report, for both the baselines and BiasEdit, the test accuracy at the checkpoint selected on the in-distribution validation set and the test accuracy at the checkpoint that yields the best performance on the bias-conflict test set.}
\centering
\tiny
\setlength{\tabcolsep}{3pt}
\renewcommand{\arraystretch}{0.85}
\resizebox{\textwidth}{!}{%
  \begin{tabular}{c|ccc|ccc|ccc|ccc}
\toprule
\multicolumn{13}{c}{\textbf{Colored MNIST}}\\
\midrule
\multirow{3}{*}{Method}
& \multicolumn{6}{c|}{\textbf{1.0\%}}
& \multicolumn{6}{c}{\textbf{5.0\%}}\\
& \multicolumn{3}{c}{ID-Val selected}
& \multicolumn{3}{c|}{Best-BC}
& \multicolumn{3}{c}{ID-Val selected}
& \multicolumn{3}{c}{Best-BC}\\
& BC & BA & \cellcolor{gray!20}Avg
& Best BC & BA (at best BC) & \cellcolor{gray!20}Avg
& BC & BA & \cellcolor{gray!20}Avg
& Best BC & BA (at best BC) & \cellcolor{gray!20}Avg \\
\midrule
Vanilla~\cite{he2016deep}
& 37.84\% & 99.78\% & \cellcolor{gray!20}68.81\%
& 40.56\% & 99.83\% & \cellcolor{gray!20}70.20\%
& 76.79\% & 99.38\% & \cellcolor{gray!20}88.09\%
& 77.33\% & 99.40\% & \cellcolor{gray!20}88.37\% \\

LfF~\cite{nam2020learning}
& 50.44\% & 94.67\% & \cellcolor{gray!20}72.56\%
& 71.04\% & 82.27\% & \cellcolor{gray!20}76.66\%
& 86.89\% & 96.22\% & \cellcolor{gray!20}91.55\%
& 88.47\% & 93.20\% & \cellcolor{gray!20}90.83\% \\

LfF+BE~\cite{lee2023revisiting}
& 34.13\% & 94.73\% & \cellcolor{gray!20}64.43\%
& 75.29\% & 81.31\% & \cellcolor{gray!20}78.30\%
& 75.18\% & 86.58\% & \cellcolor{gray!20}80.88\%
& 87.97\% & 84.19\% & \cellcolor{gray!20}86.08\% \\

SoftCon~\cite{hong2021unbiased}
& 34.18\% & 99.80\% & \cellcolor{gray!20}66.99\%
& 36.93\% & 99.50\% & \cellcolor{gray!20}68.21\%
& 68.66\% & 98.87\% & \cellcolor{gray!20}83.77\%
& 70.83\% & 98.54\% & \cellcolor{gray!20}84.69\% \\

DFA~\cite{lee2021learning}
& 25.17\% & 97.10\% & \cellcolor{gray!20}61.13\%
& 72.48\% & 88.54\% & \cellcolor{gray!20}80.51\%
& 85.71\% & 96.56\% & \cellcolor{gray!20}91.14\%
& 87.25\% & 94.85\% & \cellcolor{gray!20}91.05\% \\

SelecMix~\cite{hwang2022selecmix}
& 73.54\% & 97.21\% & \cellcolor{gray!20}85.38\%
& 78.85\% & 93.24\% & \cellcolor{gray!20}86.05\%
& 89.46\% & 94.00\% & \cellcolor{gray!20}91.73\%
& 90.16\% & 90.65\% & \cellcolor{gray!20}90.41\% \\

BiasAdv~\cite{lim2023biasadv}
& 40.40\% & 99.90\% & \cellcolor{gray!20}70.15\%
& 43.20\% & 99.90\% & \cellcolor{gray!20}71.55\%
& 78.31\% & 99.47\% & \cellcolor{gray!20}88.89\%
& 78.99\% & 99.53\% & \cellcolor{gray!20}89.26\% \\

DeNetDM~\cite{sreelatha2024denetdm}
& 44.94\% & 100.00\% & \cellcolor{gray!20}72.47\%
& 46.54\% & 1.00\% & \cellcolor{gray!20}23.77\%
& 80.72\% & 99.70\% & \cellcolor{gray!20}90.21\%
& 82.64\% & 99.76\% & \cellcolor{gray!20}91.20\% \\

\midrule
\textbf{BiasEdit (ours)}
& 92.73\% & 96.42\% & \cellcolor{gray!20}\textbf{94.58\%}
& 93.52\% & 96.42\% & \cellcolor{gray!20}\textbf{94.97\%}
& 91.72\% & 96.98\% & \cellcolor{gray!20}\textbf{94.35\%}
& 92.36\% & 97.20\% & \cellcolor{gray!20}\textbf{94.78\%} \\
\midrule
\bottomrule

\toprule
\multicolumn{13}{c}{\textbf{BFFHQ}}\\
\midrule
\multirow{3}{*}{Method}
& \multicolumn{6}{c|}{\textbf{1.0\%}}
& \multicolumn{6}{c}{\textbf{5.0\%}}\\
& \multicolumn{3}{c}{ID-Val selected}
& \multicolumn{3}{c|}{Best-BC}
& \multicolumn{3}{c}{ID-Val selected}
& \multicolumn{3}{c}{Best-BC}\\
& BC & BA & \cellcolor{gray!20}Avg
& Best BC & BA (at best BC) & \cellcolor{gray!20}Avg
& BC & BA & \cellcolor{gray!20}Avg
& Best BC & BA (at best BC) & \cellcolor{gray!20}Avg \\
\midrule
Vanilla~\cite{he2016deep}
& 54.54\% & 99.16\% & \cellcolor{gray!20}76.85\%
& 58.76\% & 98.80\% & \cellcolor{gray!20}78.78\%
& 76.88\% & 99.16\% & \cellcolor{gray!20}88.02\%
& 80.84\% & 98.73\% & \cellcolor{gray!20}89.79\% \\

LfF~\cite{nam2020learning}
& 54.64\% & 91.56\% & \cellcolor{gray!20}73.10\%
& 65.72\% & 77.00\% & \cellcolor{gray!20}71.36\%
& 72.60\% & 89.76\% & \cellcolor{gray!20}81.18\%
& 78.96\% & 82.68\% & \cellcolor{gray!20}80.82\% \\

LfF+BE~\cite{lee2023revisiting}
& 55.60\% & 84.93\% & \cellcolor{gray!20}70.27\%
& 73.53\% & 76.73\% & \cellcolor{gray!20}75.13\%
& 68.06\% & 77.06\% & \cellcolor{gray!20}72.56\%
& 84.93\% & 69.60\% & \cellcolor{gray!20}77.27\% \\

SoftCon~\cite{hong2021unbiased}
& 59.40\% & 99.30\% & \cellcolor{gray!20}79.35\%
& 62.10\% & 99.00\% & \cellcolor{gray!20}80.55\%
& 78.80\% & 99.40\% & \cellcolor{gray!20}89.10\%
& 83.80\% & 98.60\% & \cellcolor{gray!20}91.20\% \\

DisEnt~\cite{lee2021learning}
& 51.48\% & 93.44\% & \cellcolor{gray!20}72.46\%
& 63.48\% & 87.74\% & \cellcolor{gray!20}75.61\%
& 73.36\% & 94.16\% & \cellcolor{gray!20}83.76\%
& 78.64\% & 87.88\% & \cellcolor{gray!20}83.26\% \\

SelecMix~\cite{hwang2022selecmix}
& 62.76\% & 79.96\% & \cellcolor{gray!20}71.36\%
& 73.56\% & 64.98\% & \cellcolor{gray!20}69.27\%
& 76.28\% & 72.84\% & \cellcolor{gray!20}74.56\%
& 82.92\% & 61.31\% & \cellcolor{gray!20}72.12\% \\

BiasAdv~\cite{lim2023biasadv}
& 53.13\% & 98.26\% & \cellcolor{gray!20}75.70\%
& 57.26\% & 96.93\% & \cellcolor{gray!20}77.10\%
& 68.13\% & 97.33\% & \cellcolor{gray!20}82.73\%
& 73.60\% & 97.26\% & \cellcolor{gray!20}85.43\% \\

DeNetDM~\cite{sreelatha2024denetdm}
& 59.53\% & 98.40\% & \cellcolor{gray!20}78.97\%
& 64.13\% & 98.00\% & \cellcolor{gray!20}81.07\%
& 70.78\% & 98.46\% & \cellcolor{gray!20}84.62\%
& 75.00\% & 98.89\% & \cellcolor{gray!20}86.95\% \\

\midrule
\textbf{BiasEdit (ours)}
& 69.40\% & 98.80\% & \cellcolor{gray!20}\textbf{84.10\%}
& 75.40\% & 97.33\% & \cellcolor{gray!20}\textbf{86.37\%}
& 82.73\% & 98.20\% & \cellcolor{gray!20}\textbf{90.47\%}
& 85.60\% & 98.00\% & \cellcolor{gray!20}\textbf{91.80\%} \\
\midrule
\bottomrule

\toprule
\multicolumn{13}{c}{\textbf{Dogs \& Cats}}\\
\midrule
\multirow{3}{*}{Method}
& \multicolumn{6}{c|}{\textbf{1.0\%}}
& \multicolumn{6}{c}{\textbf{5.0\%}}\\
& \multicolumn{3}{c}{ID-Val selected}
& \multicolumn{3}{c|}{Best-BC}
& \multicolumn{3}{c}{ID-Val selected}
& \multicolumn{3}{c}{Best-BC}\\
& BC & BA & \cellcolor{gray!20}Avg
& Best BC & BA (at best BC) & \cellcolor{gray!20}Avg
& BC & BA & \cellcolor{gray!20}Avg
& Best BC & BA (at best BC) & \cellcolor{gray!20}Avg \\
\midrule
Vanilla~\cite{he2016deep}
& 27.28\% & 96.75\% & \cellcolor{gray!20}62.01\%
& 37.90\% & 93.39\% & \cellcolor{gray!20}65.65\%
& 56.55\% & 96.00\% & \cellcolor{gray!20}76.28\%
& 59.85\% & 95.52\% & \cellcolor{gray!20}77.69\% \\

LfF~\cite{nam2020learning}
& 46.10\% & 74.58\% & \cellcolor{gray!20}60.34\%
& 70.75\% & 52.05\% & \cellcolor{gray!20}61.40\%
& 48.28\% & 67.70\% & \cellcolor{gray!20}57.99\%
& 82.10\% & 32.82\% & \cellcolor{gray!20}57.46\% \\

LfF+BE~\cite{lee2023revisiting}
& 58.04\% & 52.08\% & \cellcolor{gray!20}55.06\%
& 83.33\% & 29.00\% & \cellcolor{gray!20}56.17\%
& 75.45\% & 45.83\% & \cellcolor{gray!20}60.64\%
& 88.16\% & 24.66\% & \cellcolor{gray!20}56.41\% \\

SoftCon~\cite{hong2021unbiased}
& 32.00\% & 97.62\% & \cellcolor{gray!20}64.81\%
& 41.62\% & 90.03\% & \cellcolor{gray!20}65.83\%
& 69.50\% & 97.38\% & \cellcolor{gray!20}83.44\%
& 73.75\% & 96.12\% & \cellcolor{gray!20}84.94\% \\

DFA~\cite{lee2021learning}
& 33.73\% & 88.00\% & \cellcolor{gray!20}60.86\%
& 65.90\% & 57.22\% & \cellcolor{gray!20}61.56\%
& 47.75\% & 83.30\% & \cellcolor{gray!20}65.53\%
& 80.93\% & 42.55\% & \cellcolor{gray!20}61.74\% \\

SelecMix~\cite{hwang2022selecmix}
& 41.05\% & 65.40\% & \cellcolor{gray!20}53.23\%
& 88.18\% & 22.77\% & \cellcolor{gray!20}55.47\%
& 40.38\% & 60.25\% & \cellcolor{gray!20}50.31\%
& 89.40\% & 21.51\% & \cellcolor{gray!20}55.46\% \\

BiasAdv~\cite{lim2023biasadv}
& 21.08\% & 94.07\% & \cellcolor{gray!20}57.58\%
& 35.54\% & 82.91\% & \cellcolor{gray!20}59.23\%
& 40.12\% & 93.33\% & \cellcolor{gray!20}66.73\%
& 51.29\% & 82.87\% & \cellcolor{gray!20}67.08\% \\

DeNetDM~\cite{sreelatha2024denetdm}
& 17.66\% & 95.16\% & \cellcolor{gray!20}56.41\%
& 29.14\% & 93.95\% & \cellcolor{gray!20}61.55\%
& 38.33\% & 94.12\% & \cellcolor{gray!20}66.23\%
& 49.91\% & 93.20\% & \cellcolor{gray!20}71.56\% \\

\midrule
\textbf{BiasEdit (ours)}
& 63.04\% & 97.08\% & \cellcolor{gray!20}\textbf{80.06\%}
& 71.79\% & 94.00\% & \cellcolor{gray!20}\textbf{82.90\%}
& 77.70\% & 97.16\% & \cellcolor{gray!20}\textbf{87.43\%}
& 83.00\% & 91.29\% & \cellcolor{gray!20}\textbf{87.15\%} \\
\midrule
\midrule
\bottomrule

\toprule
\multicolumn{13}{c}{\textbf{Waterbirds}}\\
\midrule
\multirow{3}{*}{Method}
& \multicolumn{6}{c|}{\textbf{1.0\%}}
& \multicolumn{6}{c}{\textbf{5.0\%}}\\
& \multicolumn{3}{c}{ID-Val selected}
& \multicolumn{3}{c|}{Best-BC}
& \multicolumn{3}{c}{ID-Val selected}
& \multicolumn{3}{c}{Best-BC}\\
& BC & BA & \cellcolor{gray!20}Avg
& Best BC & BA (at best BC) & \cellcolor{gray!20}Avg
& BC & BA & \cellcolor{gray!20}Avg
& Best BC & BA (at best BC) & \cellcolor{gray!20}Avg \\
\midrule
Vanilla~\cite{he2016deep}
& 27.73\% & 95.14\% & \cellcolor{gray!20}61.44\%
& 44.63\% & 92.39\% & \cellcolor{gray!20}68.51\%
& 46.00\% & 94.05\% & \cellcolor{gray!20}70.02\%
& 60.19\% & 91.77\% & \cellcolor{gray!20}75.98\% \\

LfF~\cite{nam2020learning}
& 58.60\% & 75.64\% & \cellcolor{gray!20}67.12\%
& 70.65\% & 47.95\% & \cellcolor{gray!20}59.30\%
& 70.64\% & 72.39\% & \cellcolor{gray!20}71.52\%
& 77.36\% & 31.40\% & \cellcolor{gray!20}54.38\% \\

LfF+BE~\cite{lee2023revisiting}
& 70.70\% & 43.19\% & \cellcolor{gray!20}56.95\%
& 80.37\% & 41.33\% & \cellcolor{gray!20}60.85\%
& 81.22\% & 34.66\% & \cellcolor{gray!20}57.94\%
& 86.23\% & 36.69\% & \cellcolor{gray!20}61.46\% \\

SoftCon~\cite{hong2021unbiased}
& 32.27\% & 94.78\% & \cellcolor{gray!20}63.53\%
& 42.45\% & 94.63\% & \cellcolor{gray!20}68.54\%
& 48.94\% & 94.51\% & \cellcolor{gray!20}71.73\%
& 57.92\% & 91.00\% & \cellcolor{gray!20}74.46\% \\

DisEnt~\cite{lee2021learning}
& 44.79\% & 89.05\% & \cellcolor{gray!20}66.92\%
& 77.13\% & 40.22\% & \cellcolor{gray!20}58.68\%
& 58.19\% & 88.13\% & \cellcolor{gray!20}73.16\%
& 77.92\% & 26.39\% & \cellcolor{gray!20}52.16\% \\

SelecMix~\cite{hwang2022selecmix}
& 49.47\% & 82.63\% & \cellcolor{gray!20}66.05\%
& 85.81\% & 33.71\% & \cellcolor{gray!20}59.76\%
& 52.67\% & 78.99\% & \cellcolor{gray!20}65.83\%
& 87.08\% & 21.17\% & \cellcolor{gray!20}54.12\% \\

BiasAdv~\cite{lim2023biasadv}
& 33.65\% & 93.61\% & \cellcolor{gray!20}63.63\%
& 49.46\% & 90.73\% & \cellcolor{gray!20}70.10\%
& 40.42\% & 92.45\% & \cellcolor{gray!20}66.44\%
& 54.79\% & 89.54\% & \cellcolor{gray!20}72.17\% \\

DeNetDM~\cite{sreelatha2024denetdm}
& 28.63\% & 94.66\% & \cellcolor{gray!20}61.65\%
& 40.19\% & 93.32\% & \cellcolor{gray!20}66.76\%
& 41.52\% & 93.78\% & \cellcolor{gray!20}67.65\%
& 56.30\% & 88.59\% & \cellcolor{gray!20}72.45\% \\

\midrule
\textbf{BiasEdit (ours)}
& 59.79\% & 90.77\% & \cellcolor{gray!20}\textbf{75.28\%}
& 74.28\% & 80.92\% & \cellcolor{gray!20}\textbf{77.60\%}
& 68.57\% & 89.90\% & \cellcolor{gray!20}\textbf{79.24\%}
& 75.98\% & 83.98\% & \cellcolor{gray!20}\textbf{79.98\%} \\
\midrule
\bottomrule
\end{tabular}%
}
\label{tab:additional_results}
\end{table*}

\end{document}